%% file: archiveRevised.tex
\documentclass{article}
\usepackage{arxiv}
\usepackage[utf8]{inputenc}
\usepackage[pdftex]{graphicx}
\usepackage{mathtools}
\usepackage{amsmath}
\usepackage{array}
\usepackage{mathrsfs}
\usepackage{amssymb}
\usepackage{booktabs}
\usepackage{multirow}
\usepackage{tabularx}
\usepackage{tabulary}
\usepackage{dcolumn}
\usepackage{threeparttable}
\usepackage{bbm}
\usepackage{color, soul}
\usepackage{blindtext}
\usepackage{setspace}
\usepackage{geometry}
\usepackage{wrapfig}
\usepackage{lscape}
\usepackage{rotating}
\usepackage{epstopdf}
\usepackage{wasysym}
\usepackage[]{algorithm}
\usepackage{algpseudocode}
\usepackage{amsfonts}
\usepackage[toc,page]{appendix}
\usepackage{siunitx}
\usepackage{gensymb}
\usepackage{changes}
\usepackage{etoolbox}
\usepackage{multirow}
\usepackage{afterpage}
\usepackage{comment}
\usepackage{pgf}
\usepackage{siunitx}
\usepackage{standalone}
\usepackage{hyperref}
\usepackage[english]{babel}
\usepackage{fancyvrb}
\usepackage[perpage]{footmisc}

\hypersetup{
	colorlinks,
	linkcolor={blue!75!black},
	citecolor={blue!75!black},
	urlcolor={blue!75!black}
}
\providecommand{\keywords}[1]
{
	\small	
	\textbf{\textit{Keywords:}} #1
}

\allowdisplaybreaks
\usepackage{tikz}
\usetikzlibrary{shapes,arrows,fit,positioning}
\usepackage[numbers,sort&compress]{natbib}

\allowdisplaybreaks 

\title{Feature Engineering and Forecasting via Derivative-free Optimization and Ensemble of Sequence-to-sequence Networks with Applications in Renewable Energy}

\author{
	Mohammad Pirhooshyaran \\
	Industrial and Systems Engineering\\
	Lehigh University\\
	\texttt{mop216@lehigh.edu} \\
	\And
	Katya Scheinberg \\
	Operations Research and Information Engineering\\
	Cornell University\\
	\texttt{ks2375@cornell.edu}
	\And 
	Lawrence V.\ Snyder \\
	Industrial and Systems Engineering\\
	Lehigh University\\
	\texttt{lvs2@lehigh.edu}
}

\begin{document}
	
	\maketitle

	\begin{abstract}
		This study introduces a framework for the forecasting, reconstruction and feature engineering of multivariate processes along with its renewable energy applications. We integrate derivative-free optimization with an ensemble of sequence-to-sequence networks and design a new resampling technique called additive resampling, which, along with Bootstrap aggregating (bagging) resampling, are applied to initialize the ensemble structure. Moreover, we explore the proposed framework performance on three renewable energy sources---wind, solar and ocean wave---and conduct several short- to long-term forecasts showing the superiority of the proposed method compared to numerous machine learning techniques. The findings indicate that the introduced method performs more accurately when the forecasting horizon becomes longer. In addition, we modify the framework for automated feature selection. The model represents a clear interpretation of the selected features. Furthermore, we investigate the effects of different environmental and marine factors on the wind speed and ocean output power, respectively, and report the selected features. Finally, we explore the online forecasting setting and illustrate that the model outperforms alternatives through different measurement errors.    
		
	\end{abstract}

\section{Introduction}

Machine learning (ML) approaches such as support vector regression (SVR) and neural networks are considered as the state-of-the-art methods for multivariate multistep forecasting in particular for renewable energy systems \cite{zhou2011fine,li2010comparing,pirhooshyaran2019multivariate}. We argue there can be two possible improvements. First of all, neural networks introduced in the past largely contain simple feed-forward or long short term memory (LSTM) structures and do not consider more advanced structures where the method can benefit the most from temporal concept in their design. Second, even though by their nature, many time series, including but not limited to renewable energy features such as wind speed or significant wave height, are intermittent and stochastic, they are periodic as well and express repetitive behavior. It is well-known for instance that the wind power generation seasonality generally accords with the energy demand distribution \cite{davy2018climate}. That is, during the winter season which the electricity consumption is normally higher, the output power produced by the wind turbines is higher as well. Hence, wind renewable energy contains repetitive patterns holistically. We aim to capture this inherent seasonality of alternative energy by introducing parallel network designs and aggregating their results. Furthermore, we introduce a resampling technique which accounts for the seasonality in the dataset.

DFO has emerged once again as a robust hyperparameter tuning technique for black-box functions in particular complex networks \cite{conn2009introduction}. Here, not only do we use DFO as a means to tune the sequence-to-sequence artificial networks, but we also utilize it to design the parallel structure. In other words, we use DFO to explore the optimal number of parallel networks for any specific task separately. The paper contributions are as follows:

\begin{itemize}
	\item We introduce ensemble of sequence-to-sequence networks capable of: 1) short to long term forecasting and 2) feature selection 3) online forecasting.
	\item We propose the Scaled DFO (SDFO) algorithm and integrate it with the parallel structure of sequence-to-sequence networks.
	\item We introduce a resampling technique which we call additive resampling. Furthermore, We utilize bootstrap aggregating resampling technique \cite{breiman1996bagging} to initialize the ensemble structure.  
\end{itemize}

The rest of the paper is organized as follows: In Section \ref{sec:lit}, we discuss a brief literature review. Section \ref{sec:model} introduces the model and resampling techniques. Section \ref{sec:exp} provides the implementation of the framework in three separate renewable energy environment and compares the performance results for several settings. We end the paper by a conclusion in section \ref{sec:conclu}.

\section{Literature Review}
\label{sec:lit}
There are several forecasting and feature selection methodologies for multivariate time series analysis. One may categorize the methods into Naive predictor, physical and meteorological (model-based), statistical, soft computing and hybrid approaches \cite{soman2010review}.

Persistence (or the naive approach) \cite{woon2014data} straightforwardly assumes that the time series elements are the same at time $t$ (the last real measurement data) and $t+\nabla t$ where $\nabla t$ is the forecasting horizon \cite{madsen2005tool} without adjusting them. Hence, this method expresses satisfactory behavior for forecasting very short steps into the future but for large horizons, the correlation coefficient between its predictions and the real measurements can even be negative \cite{madsen2005standardizing}.

Physical and meteorological approaches use explicit complex mathematical models to forecast the unknown. In other words, these frameworks utilize the data to construct a model and then predict the future outcome with that model. Therefore, they are more robust in cases of moderate- to long-term forecasting. Numerical weather prediction (NWP) is an umbrella term encompassing these techniques for wind models \cite{coiffier2011fundamentals}. These models sometimes suffer from lack of generalization. That is, the models produced are not applicable for many cases.

Both statistical and machine learning methods predict future outcomes solely based on historical data. Therefore, one may call them model-free approaches. Statistical studies \cite{kavasseri2009day} mainly apply time series analyses, such as Auto Regressive Integrated Moving Average (ARIMA) for prediction. Recently, ARIMA predictions have been combined with other methods, resulting in hybrid frameworks. Repeated wavelet transform \cite{singh2019repeated} and Logarithmic Generalized Autoregression (LGARCH) \cite{tian2018wind} are two examples of such successful combinations.

Soft computing and machine learning approaches in the multivariate forecasting content include fuzzy logic (FL) inference, support vector machines (SVM), extreme learning machines (ELM) and neural networks (NN) \cite{khosravi2018prediction}. Most of these concepts will be discussed in Section \ref{Alt-Alg} and are compared with our proposed method. Neural networks in the wind literature are typically used in hybrid methods. For instance, adaptive neuro-fuzzy systems are employed by \cite{moreno2018wind,khosravi2018prediction} to estimate wind power and speed. However, when we exclude the networks and focus on their structure, they are not generally the state-of-the-art structures that have been used in other machine learning disciplines. There are few exceptions though. For instance, \cite{chitsazan2019wind} introduces two novel echo state networks for speed and wind forecasting.

Here, we investigate ensemble of sequence-to-sequence \cite{sutskever2014sequence} networks. Recurrent networks such as Long Short term Memory networks (LSTMs), which contain temporal properties, are the state of the art for many time-dependent machine learning tasks such as Handwritten/Speech Recognition \cite{xiong2018microsoft,zhang2018drawing,mobiny2018text}, Disease Prediction \cite{lipton2015learning,mobiny2017lung}, and music composing \cite{choi2016text}. Sequence-to-sequence networks \cite{sutskever2014sequence} consist of two separate recurrent sections called encoder and decoder, which makes the network not only experience stability, but adaptability as well, towards changeable input/output sizes.

DFO approaches, along with Bayesian Optimization (BO), are among the most well-known methods for hyperparameter tuning in machine learning studies \cite{ghanbari2017black}. DFO is designed to optimize complex functions in which function evaluations are computationally expensive and one may not have sufficient function value samples as well as times when objective function derivatives with respect to a selected parameter are unattainable or even more costly \cite{conn2009introduction}. First, we integrate DFO with an ensemble of networks because function evaluations are computationally expensive. In addition, we utilize DFO to select the ensemble design.


\section{Model}
\label{sec:model}

\subsection{A Simple Case Study to Justify the Parallel Method}
Before explaining the parallel sequence-to-sequence structure and additive resampling technique, we present a practical justification of any parallel structure by considering a very simple case study in wind energy systems. By their nature, wind features are intermittent and stochastic. We conduct a study first to analyze the existence of seasonality and, further to emphasize the fact that wind feature reconstruction and forecasting are achievable. We use an open-source dataset\footnote{From the site: \href{http://sites.ieee.org/pes-iss/data-sets/}{Power \& Energy Society Open Data Sets}; under the section Wind Based Generation; GECAD wind speed.} of the Research Group on Intelligent Engineering and Computing for Advanced Innovation and Development (GECAD\footnote{\href{http://www.gecad.isep.ipp.pt/GECAD/Pages/Presentation/Home.aspx}{www.gecad.isep.ipp.pt}}) \cite{silva2013energy,pinto2014short,ramos2013short}. To do so, a naive decomposition approach is applied through a convolution filter to the observed data, and the smoothed average is returned with the help of the \texttt{statsmodels} Python package.

\begin{figure}
	\hspace*{-2cm}
	\scalebox{0.4}{\input{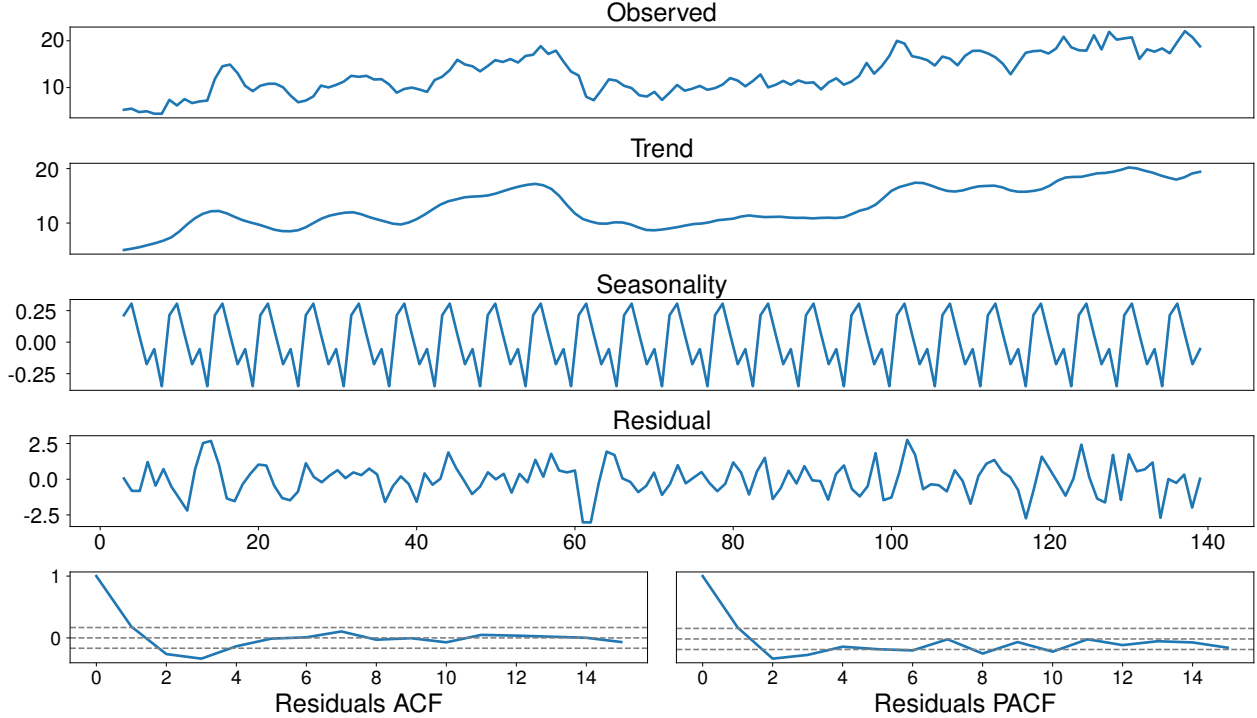}}
	\caption{Wind speed analysis of GECAD dataset (01-01-2011)}
	\label{fig:000}
\end{figure}

Figure \ref{fig:000} illustrates January 1st, 2011 observed wind data, along with its identified trend, seasonality and the remaining residual. The data has 10-minute resolution, which results in 142 daily data points. As seen in Figure \ref{fig:000}, a strong seasonality exists in the Observed data. Further, we validate the results by conducting a stationarity (Adfuller) test on residuals to monitor whether residuals behave as a stationary time series. Table \ref{tab:1000} demonstrates the results. The lag value has been selected to optimize the Akaike Information Criteria (AIC). The test statistic is considerably below the critical values for two different confidence levels, 1\% and 5\%. Hence, there is no evidence to reject the null hypothesis. In the end, we investigate the AutoCorrelation Function (ACF) and Partial AutoCorrelation Function (PACF) measures on the residuals, which are shown in the last row of Figure \ref{fig:000}. In time series analysis, one would consider a time series as white noise at a certain confidence level if there is no positive index in its ACF or PACF plots which has a value greater than that level. In this case, the series has been generated out of pure stochasticity and usually there would be no further analysis over it. In contrast, if the ACF or PACF of a time series have values greater than a confidence level, it suggests data auto-correlation and more complex structures to be explored further. Wind speed residual series, as illustrated in Figure \ref{fig:000}, can be considered in the latter category for the 0.95 confidence level. Therefore, not only does wind speed data express seasonality, but it holds extra hidden relations, as well, which convinced us to try ensemble structures.\\

\begin{table}
	\centering
	\caption{ Stationarity Adfuller test on residuals of wind speed data of GECAD dataset (01-01-2011)}
	\label{tab:1000}
	\begin{tabular}{ll}\hline
		Test statistic                   & -6.41      \\ \hline
		P-value                          & 1.841e-08  \\ \hline
		Lag                          & 9          \\ \hline
		\multirow{2}{*}{Test critical values} & 1 \%: -3.48 \\
		& 5 \%: -2.88 \\ \hline
	\end{tabular}
\end{table}

\subsection{Ensemble of Sequence-to-Sequence Networks}

\begin{figure}[]
	\hspace*{4cm}
	\includegraphics[width = 8cm]{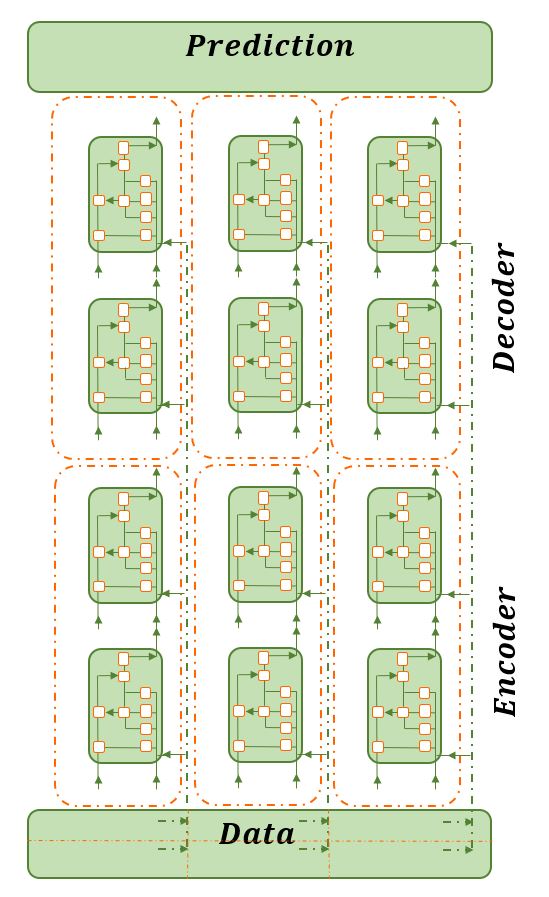}
	\caption{Ensemble of sequence-to-sequence networks}
	\label{fig:1}
\end{figure}

Here, we propose an ensemble of sequence-to-sequence networks integrated with SDFO-TR for forecasting and feature engineering of multivariate processes. Each sequence-to-sequence network is designed to approximate conditional distribution of $ P([b_{T+T'}]|[a_1,a_2,...,a_T])$ when provided with $[a_1,a_2,...,a_T]$ as input series to evaluate the output $[b_{T+T'}]$. $T$ and $T'$ are the network temporal recursion and the future horizon of the problem respectively. Figure \ref{fig:1} illustrates the ensemble structure and the details of each part.

LSTM cells are used for each sequence. LSTM cells include four gates named input, state, output and forget. We optimize the network(s) through Backpropagation with ADAM optimizer. For more details on the LSTM, sequence-to-sequence cells one may see Appendices \ref{appendix_A}. We use the Ensemble Factor (EF) term for the number of independent sequence-to-sequence networks. Figure \ref{fig:1} illustrates the ensemble structure. EF requires more study to be specified. One may argue that this factor complies with historical data seasonality. For example, for monthly data with yearly seasonality, $1, 2, 4, 6, 12$ are logical numbers to investigate. But we require an automated way to identify the optimal EF. To systematically choose the value for EF, we integrate SDFO-TR into the design of the structure.

\subsection{Resampling}
Resampling is defined as constructing a number of samples from a dataset and refitting a specific model on each sample, hoping to perceive more accurate knowledge about the model. We use two methodologies, as follows:

\begin{itemize}
	\item Additive resampling:\\
	To tackle the seasonality of the wind data, we divide the train and test datasets into a number of subsections equal to the EF. Then, we feed the whole train dataset plus a subsection of the train dataset into an independent network and evaluate a corresponding test subsection on that network. This division may result in discovering new functional relations within the subsection while they were hidden in the mixture. This method can be relatively expensive in terms of computations since each new epoch contains more samples. In the next subsection we discuss the fact that the DFO optimizes the EF number as one of its inputs. Hence, the number of the divisions for additive resampling depends on the dataset and the feature under study. We aim to reach the maximum potential of the parallel structure by using Additive resampling by optimizing the EF factor and directly using it in training and testing the networks.
	\item Bootstrap aggregating (bagging):\\
	We draw a number of samples with replacement and equal in size to the original dataset. For a large dataset, approximately $1-\dfrac{1}{e}$ fraction of any sample would be unique and the rest would be replicates \cite{buhlmann2002analyzing}. Bagging is a well-known technique for statistical resampling \cite{breiman1996bagging}.
\end{itemize}
A graphic explanation of the above techniques is shown in Figure \ref{framework}. 

\afterpage{
	\thispagestyle{empty}
	\begin{figure}[htbp]
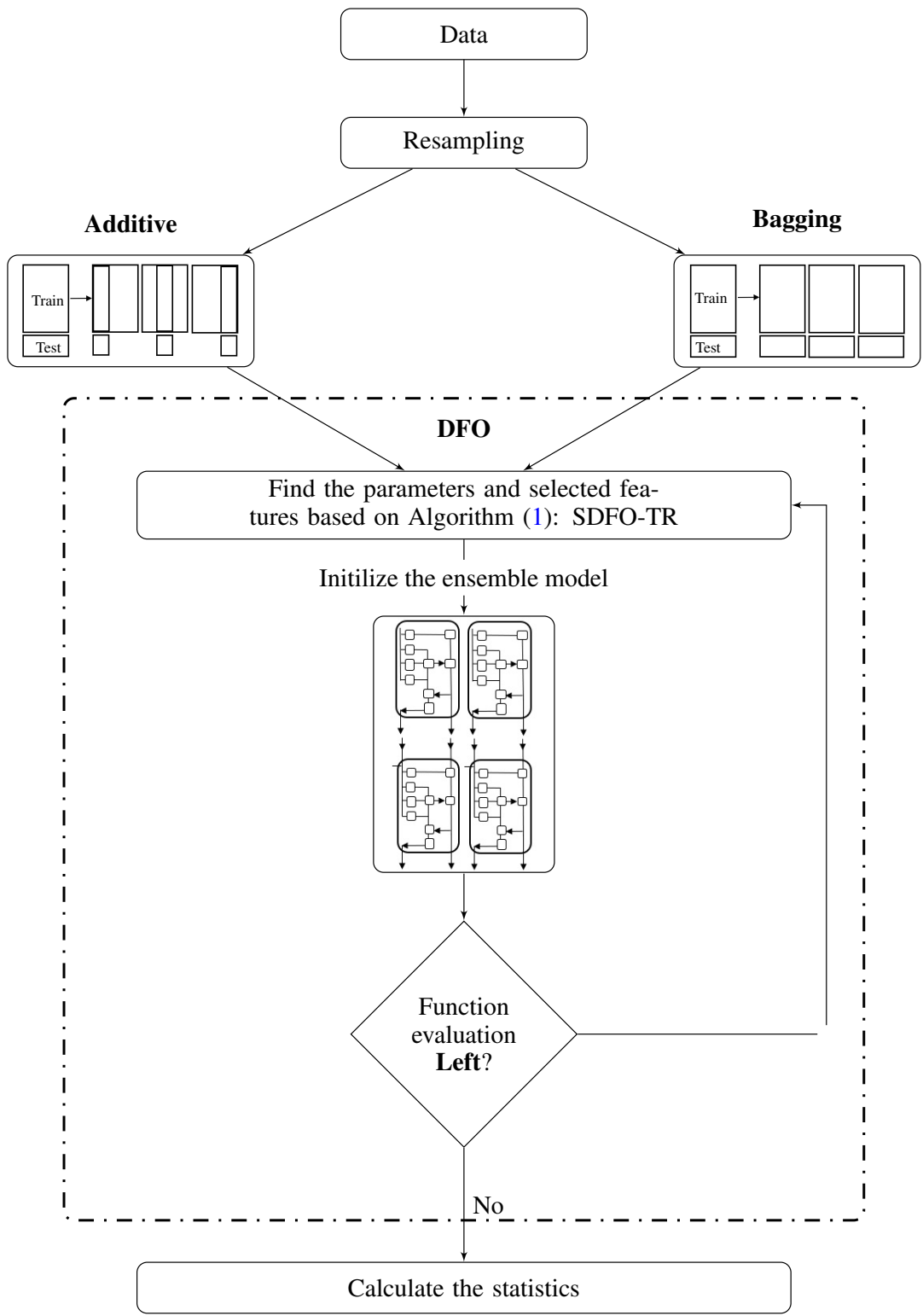

		\thispagestyle{empty}	
		\includestandalone[width = 15cm]{framework}
		\caption{Proposed Framework}
		\label{framework}
	\end{figure} 
	\clearpage
}
\subsection{Scaled Trust Region Derivative-Free Optimization (SDFO-TR)}

We modify a trust region version of DFO (DFO-TR). Compared to Bayesian optimization, DFOs are known to be time efficient and accurate, but the main deficiency is that in most studied DFO algorithms, the parameters are required to be purely continuous (their domain should be $\mathbb{R}$). Methods using DFO for integer variables basically boil down to enumeration or heuristics plus DFO for pure continuous variables. Here, we present scaled DFO-TR method capable of optimizing integer parameters. Algorithm \ref{alg:1} indicates in detail the scaled DFO-TR approach. This algorithm is based on \cite{ghanbari2017black,conn2009introduction}, with the following modifications:

\begin{algorithm}
	\caption{SDFO-TR}
	\label{alg:1}
	\begin{small}
		\begin{algorithmic}[1]
			\State \textbf{Input: }Budget $\mathcal{L}, $ $z_0 >0$, $0<\gamma_1 <1<\gamma_2$, $0<\eta_0 < \eta_1 <1$, $\theta >1$, penalty coefficient $b$, upper bound on the function value $c$ and $[l^i,u^i] \ i=1,2,..., P$.
			\State Initialize $x_0 =(x_0^1,...,x_0^P)$ where   $x^i_0 \in [l^i,u^i] \ \forall i=1,2,..., P$.
			\State Determine set of interpolation points $X \subseteq \mathcal{B}(x_0,z_0)$ to construct the model and evaluate $f(x), \forall x \in X$.
			\State $x_0 \leftarrow \text{argmin}_{x \in X} f(x) $.
			\For{$k=0,1,2,...,\mathcal{L}$ }{	
				\If{$x_k^i \in [l^i,u^i]$ $ \ \forall  i=1,2,..., P$} {
					\State Discard all points $x \in X$ where $ \|x-x_k \| \geq \theta z_k$.
					\State $M_k(x_k)\leftarrow f_k + g_k^T(x-x_k)+\dfrac{1}{2}(x-x_k)^TH_k(x-x_k) $.
					\State $\bar{x}_k =\text{argmin}_{x \in \mathcal{B}(x_k,z_k)} M_k(x)$.
					\State $\rho_k=\dfrac{f(x_k)- f(\bar{x}_k)}{M_k(x_k)-M_k(\bar{x}_k)}$.
					\If {$|X| < \dfrac{(P+1)(P+2)}{2}$}
					\State Append $\bar{x}_k$ to the set of interpolation points $X$.
					\ElsIf { $\big(\rho_k \geq \eta_0$ or $\|x - x_k \| < \max_{x \in X} \| x - x_k \| \big)$} 
					\State Substitute the point $ \max_{x \in X} \| x - x_k \|$ with $\bar{x}_k$ in the set $X$. 
					\EndIf			         
					\If {$\rho_k \geq \eta_1$}
					\State $x_{k+1} \leftarrow \bar{x}_k$ and $z_k \leftarrow \gamma_2 z_k$.
					\ElsIf{ $\rho_k < \eta_0$}
					{\State $x_{k+1} \leftarrow x_k$
						\If {$|X|> (P+1)$}
						\State $z_{k+1} \leftarrow \gamma_1 z_k$ 
						\Else 
						\State $z_{k+1} \leftarrow {z_k}$
						\EndIf}
					\ElsIf {   $\eta_0 \leq \rho_k < \eta_1$}
					{\State $x_{k+1} \leftarrow \bar{x}_k$
						\State $z_{k+1} \leftarrow {z_k}$ \EndIf}}	
				\State \Return  $ M_k(x_k)$
				\Else
				\State	\Return$c+b \sum_{i \in I^*} \min\{ |x^i-l^i|, |x^i-u^i|  \}$ where set $I^*$ contains indices for which $x^i \notin [l^i,u^i]$
				\EndIf
				
				\EndFor	}	
			
		\end{algorithmic}
	\end{small}
\end{algorithm}

1) The model is designed to minimize a black-box function in which the variables are within specified intervals instead of taking the whole $\mathbb{R}$ space. To achieve this, we further assume that the black-box loss function has an upper bound. Then for any parameter violating its interval, we consider the upper bound plus a penalty cost linearly related to the violation distance as the ultimate value of the function. This value would be considerably larger than the function value when all variables are inside their feasible regions. Therefore, the model has a logical incentive for not going out of the parameters' feasible intervals. The observations are completely in support of this technique. At the beginning, the model explores through the parameters' space randomly, and after some initial iterations, it mainly remains within their acceptable intervals. The mean arctangent absolute percentage error (MAAPE) has the known upper bound of $\dfrac{\pi}{2}$ (see \ref{MAAPE}). So, we mainly use this as the DFO loss function.

2) We further scale the intervals to $[0,1]$. Theoretically, scaling is not required because the trust region radius would handle different magnitudes for different parameters. However, scaling assists DFO in maintaining a good margin within acceptable intervals.

3) We bring the complexity of dealing with integer variables into the objective value. We divide the $[0,1]$ intervals into sub-intervals for which the objective value would remain unchanged. To fully grasp the idea, let's assume that the proposed DFO optimizes a sequence-to-sequence ensemble considering the EF as its only input, and let's further assume that EF $\in \{1,2,4,6\} $. We use MAAPE as the DFO loss function. Then, we can say algorithm \ref{alg:1} minimizes the following piecewise black-box function:

\[\text{DFO}(x_{ef})=\begin{cases} 
\dfrac{\pi}{2} -2x_{ef} & x_{ef} < 0 \\
S(x_{ef}=1) & 0\leq x_{ef} < 0.25 \\
S(x_{ef}=2) & 0.25\leq x_{ef} < 0.5 \\
S(x_{ef}=4)& 0.5\leq x_{ef} < 0.75 \\
S(x_{ef}=6)& 0.75\leq x_{ef} < 1 \\
\dfrac{\pi}{2} +2(x_{ef}-1) & x_{ef} > 1, \\
\end{cases}
\] 
where $x_{ef} \in \mathbb{R}$ represents EF as the input variable, and $S(x_{ef}=i$) is the objective value of the parallel structure when we have $i$ independent sequence-to-sequence networks, $i \in \{1,2,4,6\}$. We considered the penalty coefficient $b=2$. Based on the scaling techniques we used in algorithm \ref{alg:1}, we call this method SDFO-TR.

As shown in lines 1-4 of Algorithm \ref{alg:1}, it starts to build an interpolation set based on some early function evaluations. Therefore, the initial value $x_0 =(x_0^1,...,x_0^P)$ plays an important role, and having an acceptable guess may increase the performance \cite{conn1997convergence}. There exist three major blocks inside the for loop:
\begin{itemize}
	\item First block (lines 6-10): The algorithm constructs a quadratic approximation model $M_k(x)$ with current set of points and then determines its minimizer over the open ball $\mathcal{B}(x_k,z_k)$ where $z_k$ is the trust region radius. The success factor
	\begin{equation}
	\rho_k=\dfrac{f(x_k)- f(\bar{x}_k)}{M_k(x_k)-M_k(\bar{x}_k)} 
	\end{equation} 
	
	is the measure of how well the model $M_k(x)$ mimics the actual black-box function. 
	
	\item Second block (lines 11-15): The algorithm updates its set of interpolation points $X$ while keeping it poised. A quadratic polynomial $M(x)$ is an interpolation of a function $f(x)$ with respect to the set $X=\{x_j | j=1,...,m\}$ if
	\begin{equation}
	M(x_j)=f(x_j), \ \forall j=1,...,m.
	\end{equation} 
	The set $X$ must remain poised with respect to the linear space of polynomials building $M(x_j)$. We refer the reader to \cite{conn2009introduction,rios2013derivative} for details on poisedness of $X$. 
	
	\item Third block (lines 16-29): The Trust Region (TR) update concept is borrowed. If the success factor $\rho > \eta_1 $ which means it is close to one, then the algorithm both accepts the step and increases the radius to $\gamma_2 z_k$, in order to explore the black-box function space more aggressively. If $\eta_0 \leq \rho < \eta_1 $, the algorithm still accepts the step but the radius remains the same. Finally, if $\rho_k < \eta_0$, then the algorithm rejects the step. The TR radius would stay the same if the quadratic approximation is based on at most $P$ evaluations, but if the approximation is based on more evaluations, it means the black-box objective function, fluctuates considerably over this particular region and the radius should be decreased.  
\end{itemize}

In the instance we mentioned earlier, we solely examined the effects of one parameter to illustrate the SDFO-TR method. In general, any integer parameter with $t$ different feasible value, results in $t+2$ sections in the DFO objective function independently. We are dealing with training epochs, mini-batch sample size, LSTM hidden size, recurrent time step, optimization learning rate, number of stacked recurrent layers and EF hyperparameters in this proposal, and all of them receive integer values. Therefore, the DFO objective function initially contains millions of sections for above hyperparameter tuning. One may argue that by increasing the integer variables, the method could soon be intractable. To answer this, we know there can be several trends inside the feasible region for a parameter; for instance, like many other studies \cite{li2014efficient}, increasing the mini-batch sample size after a limit would not help the network performance. There are studies showing that DFO may escape flat local minima and clear basins \cite{ghanbari2017black,more2009benchmarking}. Therefore, DFO avoids those points and many of the sections related to them. This is why we introduced the complexity into the function at the first place and let the DFO decides.


\section{Experiments}
\label{sec:exp}
In this section we evaluate the performance of the model for several tasks. First, we focus on the pure forecasting of wind speed, direction and power based on historical data of the same feature. Second, we aim to analyze the effects of wind speed on the actual output power produced by turbines.       
\subsection{Measures of Difference}
Considering $b_1, \dots ,b_{n}$ and $\hat{b}_1,\dots,\hat{b}_{n}$ as real and predicted measurements respectively, we mainly use the following measurement errors:

{\bf NRMSE}:
\begin{equation}
\text{nrmse}=\dfrac{100 \sqrt{ \dfrac{1}{n} \sum_{i=1}^{n} (b_i-\hat{b}_i) }}{R} 
\end{equation}
where $R$ is the range of the real measurements. Root Mean Squared Error (RMSE) is one of the most frequent measurement errors, but because it is scale dependent and we use this measure for different datasets, it is more logical to use its normalized version.

{\bf MAE}:
\begin{equation}
\text{MAE}= \dfrac{\sum_{i=1}^{n} |b_i-\hat{b}_i|}{n}   
\end{equation}

Mean Absolute Error accounts for the average absolute difference between prediction and measurements.

\label{MAAPE}
{\bf MAAPE}:

Mean Absolute Percentage Error (MAPE) expresses measurement absolute accuracy as a percentage. Mean Arctangent Absolute Percentage Error (MAAPE) intrinsically maintains the MAPE scale independency and intuitive interpretability while alleviating the drawback of dividing by zero.   
\begin{equation}
\frac{1}{n} \sum_{i=1}^{n} \text{arct} \left(\left|\dfrac{b_i -\hat{b}_i}{b_i}\right| \right).
\end{equation}

\subsection{Alternative Algorithms}
\label{Alt-Alg}

In this subsection, we summarize the other approaches in the literature which we compare the proposed algorithm with.\\
\textbf{Seasonal auto regressive integrated moving average (SARIMA)}: Auto regressive moving average models \cite{box2015time,pirhooshyaran2017optimization,pirhooshyaran2015double} decompose the series into moving average and auto regressive ones. ``Integrated'' means the model uses a differencing process to reach stationarity considering a specific confidence level. A time series expresses seasonality when its expected value changes in a recurring manner. Seasonal ARIMA accounts for cases when the data exhibits seasonality. SARIMA models can be written as ARIMA$(p,d,q)(P,D,Q)_m$ where $p$ and $P$ are the time lags in the auto-regression for the trend and seasonal part, $d$ and $D$ refer to the differencing degree for trend and seasonal part, $q$ and $Q$ indicate the moving average order for both, and $m$ specifies how many periods are in each season. For time series $x_0, x_1,\cdots, x_T$, we define lag operator $L$ as: $L (x_{t}) = x_{t - 1}, t=1,\cdots,T$. Then, one may implicitly express Seasonal ARIMA$(p,d,q)(P,D,Q)_m$ in a general form as follows:
$$(1 - \sum_{i=1}^{p}\phi_{i}L^i)(1 - \sum_{i=1}^{p}\Phi_{i}L^{mi})(1 - L)^{d}(1 - L^{m})^d x_{t} = (1 + \sum_{i=1}^{q}\theta_{i}L^i)(1 + \sum_{i=1}^{q}\Theta_{i}L^{mi})\epsilon_{t}$$
where $\epsilon_t, \forall t=0,1,\cdots,T$ are zero-mean white noises, $\phi_i$ and $ \Phi_i, \forall i=1,2,\cdots,p$, are trend and seasonal auto-regression coefficients, respectively; and $ \theta_i $ and $ \Theta_i, \forall i=1,2,\cdots,q$, are trend and seasonal moving average coefficients, respectively. It is clear that satisfactory performance of seasonal ARIMA depends on how accurately we choose its parameters. We compare our proposed method with SARIMA only for the online settings.

\textbf{Extreme Learning Machine (ELM)}: ELM approaches \cite{broomhead1988radial,pao1989adaptive,huang2006extreme} aim to generalize the learning procedure of feed-forward neural networks by randomly assigning parameters of hidden layer neurons. Input-hidden or hidden-hidden weights may remain unchanged through the learning process or simply accept their parents' values in multi-layered ELM. This sharply distinguishes ELM from networks trained based on backpropagation. In addition, unlike the recursive weight updating approach used in backpropagation, one can update the input-hidden or hidden-hidden neuron weights at once in ELM, which results in several orders of magnitude faster performance time \cite{miche2010op}. Here, we mainly focus on classic ELM and its improvement called fully online sequential-extreme learning machine \cite{park2017online,wong2014adaptive} (FOSELM) for the online setting comparisons.

\textbf{Support Vector Regression}: Unlike typical regression, which tries to minimize a least squared error, SVR seeks to construct a hyperplane for which the marginal error is minimized \cite{drucker1997support}. SVR utilizes functional relations called kernels to map interconnected input data into a higher dimension where the data is (almost) linearly separable.  In SVR we have an $\epsilon$ margin of tolerance. In other words, two boundary hyperplanes exist at $\epsilon$ distance of the target hyperplane to contain most of the points, and the outliers are only allowed with a penalty cost added to the objective function. The penalty term can be looked at as a regularizer which assists the model to provide more general solutions and prevents overfitting and the most common type of them is an L2-norm. The data points on the two marginal hyperplanes that are responsible for their construction are called the support vectors. We refer the reader to MLSVM \cite{sadrfaridpour2016algebraic}, which leverages hierarchical learning SVM, and \cite{azadeh2015unique} a novel SVM case study for more details on support vectors. Here, with the help of  \texttt{LIBSVM} Python package \cite{CC01a} we implement the $\epsilon$-SVR and $\nu$-SVR methods.

\textbf{Random Forrest (RF)}: RF is a meta predictor algorithm which includes an ensemble of Classification and Regression Trees (CART)\cite{breiman2001random}. CARTs are known to be easily interpretable, and each layer of their models contain Boolean logic. On the other hand, CARTs suffer from a generalization point of view. Because of their binary boundaries, they can easily overfit the train set. RF models improve the generalization performance of CARTs to a great extent. RF utilizes the bootstrap aggregating concept to focus on different parts of the dataset for every decision tree and then uses averaging techniques for prediction similar to our introduced method. 

\subsection{Wind Renewable Energy}
As an alternative, wind energy offers a secure, competitive and sustainable option, releasing almost no greenhouse gases as well as consuming no water during its production. Dense wind farms across the globe are generating considerable amounts of electric power. The inherently time-varying nature of wind energy necessitates the development of a systematic framework to construct and forecast future wind characteristics \cite{notton2018intermittent}. Such a framework greatly enhances the flexibility of power systems to cope with irregular wind power patterns integrated in their portfolios. Therefore in the following sections, we investigate the performance of the proposed method for several tasks related to wind datasets.

\subsubsection{Forecasting}

In this part, we focus on very short- to long-term forecasting into the future considering high-resolution wind datasets. To this end, first we use an open-source national renewable energy laboratory (NREL) dataset for wind speed and wind direction forecasting \footnote{The dataset can be found at \href{http://sites.ieee.org/pes-iss/data-sets/}{Power \& Energy Society Open Data Sets (http://sites.ieee.org/pes-iss/data-sets/)}; under Weather Data section; NREL wind speed, 2010.}. Table \ref{tab:2} indicates the details of the measurement station which this data is obtained. The time period between two consecutive measurements is 5 minutes. We consider only the first ten days of 2010 (or exactly $12 \times 24 \times 10= 2880$ data points). We divide the data into 60 percent training, 20 percent validation and 20 percent testing for all methods. The design of the structure is conducted through training and then tested through validation to explore the best parameters' results. None of the methods see the test set, and further, we have not touched the real dataset measurements in any way such as interpolation or normalization.

\begin{table}
	\centering
	\caption{ National Renewable Energy Laboratory Station }
	\label{tab:2}
	\begin{tabular}{ll}\hline
		Latitude                   &   \ang{39;54;38.34} North;     \\ \hline 
		Longitude                  & \ang{105; 14; 5.28} West  \\ \hline
		Sampling period                          & 5 minutes  \\ \hline
		Elevation                           &  1855 m          \\ \hline
	\end{tabular}
\end{table}

Table \ref{tab:forecastSpeed} demonstrates the results for several wind speed forecasting horizons. In the proposed method columns, we put those Ensemble Factors corresponding to the best performance in the parentheses. One can observe: 
\begin{itemize}
	\item SVR and ensemble structure achieve more valid and precise errors compared to the other two methods.
	\item Additive resampling works marginally better than bagging one.
	\item The proposed method performs superior with the increasing of the forecasting horizon. In other words, some of the alternatives such as SVR reach smaller errors for one- or six-step-ahead estimations, but for medium- to long-term forecasting, we observe satisfactory improvement by sequence networks, particularly considering A-Sampling. We argue this can be due to the structure of the proposed method, which can discover hidden relations in deeper forecasting horizons. 
	\item Huge jumps occur in the error values by going from forecasting one step into the future to six steps. But this is not the case for deeper horizons. We even observe that some methods reach better errors for 48 steps compared to 36 steps, or 72 steps compared to 48 steps.        
\end{itemize}   

\begin{table}
	\centering
	\caption{ MAAPE, NRMSE and MAE comparison for wind speed}
	\label{tab:forecastSpeed}
	\setlength\tabcolsep{3pt}
	\hspace*{-1cm}
	
	\begin{tabular}{llllllllllll}
		MAAPE &            &  &        &  &            &            &  &        &  &                     &               \\
		&            &  & ELM    &  & SVR        &            &  & RF     &  & Proposed Method     &               \\ \cline{4-4} \cline{6-7} \cline{9-9} \cline{11-12} 
		& Time Steps &  &        &  & $\epsilon$-SVR     & $\nu$-SVR     &  &        &  & A-Sampling (EF)          & B-Sampling (EF)    \\ \cline{2-12} 
		& 1          &  & 0.153  &  & 0.151      & 0.151      &  & 0.179  &  & 0.149 (6)           & 0.154 (4)     \\
		& 6          &  & 0.335  &  & 0.325      & 0.326      &  & 0.342  &  & 0.316 (2)           & 0.320 (2)     \\
		& 12         &  & 0.415  &  & 0.401      & 0.402      &  & 0.417  &  & 0.380 (1)           & 0.397 (1)     \\
		& 24         &  & 0.478  &  & 0.447      & 0.447      &  & 0.499  &  & 0.430 (1)           & 0.447 (1)     \\
		& 36         &  & 0.503  &  & 0.465      & 0.466      &  & 0.512  &  & 0.448 (4)           & 0.477 (2)     \\
		& 48         &  & 0.500  &  & 0.461      & 0.459      &  & 0.493  &  & 0.435 (1)           & 0.468 (1)     \\
		& 72         &  & 0.519  &  & 0.459      & 0.459      &  & 0.517  &  & 0.446 (1)           & 0.487 (1)     \\ \cline{2-2} \cline{4-12} 
		&            &  &        &  &            &            &  &        &  &                     &               \\
		NRMSE &            &  &        &  &            &            &  &        &  &                     &               \\
		&            &  & ELM    &  & \multicolumn{2}{l}{SVR} &  & RF     &  & \multicolumn{2}{l}{Proposed Method} \\ \cline{4-4} \cline{6-7} \cline{9-9} \cline{11-12} 
		& Time Steps &  &        &  & $\epsilon$-SVR     & $\nu$-SVR     &  &        &  & A-Sampling          & B-Sampling    \\ \cline{2-12} 
		& 1          &  & 7.520  &  & 5.895      & 5.896      &  & 6.925  &  & 5.966               & 5.910         \\
		& 6          &  & 15.867 &  & 14.442     & 14.4284    &  & 16.328 &  & 14.392              & 14.203        \\
		& 12         &  & 20.146 &  & 17.761     & 17.7404    &  & 19.780 &  & 18.678              & 17.942        \\
		& 24         &  & 23.894 &  & 19.596     & 19.715     &  & 24.109 &  & 18.882              & 20.599        \\
		& 36         &  & 22.930 &  & 21.041     & 21.103     &  & 25.988 &  & 20.641              & 21.731        \\
		& 48         &  & 22.809 &  & 19.660     & 19.620     &  & 24.180 &  & 18.284              & 20.474        \\
		& 72         &  & 22.704 &  & 17.629     & 17.534     &  & 23.160 &  & 17.123              & 18.787        \\ \cline{2-12} 
		&            &  &        &  &            &            &  &        &  &                     &               \\
		MAE   &            &  &        &  &            &            &  &        &  &                     &               \\
		&            &  & ELM    &  & \multicolumn{2}{l}{SVR} &  & RF     &  & \multicolumn{2}{l}{Proposed Method} \\ \cline{4-4} \cline{6-7} \cline{9-9} \cline{11-12} 
		& Time Steps &  &        &  & $\epsilon$-SVR     & $\nu$-SVR     &  &        &  & A-Sampling          & B-Sampling    \\ \cline{2-2} \cline{4-7} \cline{9-12} 
		& 1          &  & 0.444  &  & 0.370      & 0.369      &  & 0.449  &  & 0.372               & 0.373         \\
		& 6          &  & 0.989  &  & 0.944      & 0.945      &  & 1.074  &  & 0.952               & 0.937         \\
		& 12         &  & 1.293  &  & 1.210      & 1.210      &  & 1.331  &  & 1.235               & 1.217         \\
		& 24         &  & 1.539  &  & 1.370      & 1.380      &  & 1.653  &  & 1.298               & 1.417         \\
		& 36         &  & 1.618  &  & 1.448      & 1.452      &  & 1.727  &  & 1.398               & 1.497         \\
		& 48         &  & 1.550  &  & 1.357      & 1.351      &  & 1.601  &  & 1.242               & 1.409         \\
		& 72         &  & 1.548  &  & 1.227      & 1.218      &  & 1.544  &  & 1.151               & 1.305         \\ \cline{2-12} 
	\end{tabular}
\end{table}

Table \ref{tab:forecastDir} indicates the results for similar forecasting horizons for wind direction on the same subsample of the NREL dataset we used before. One can observe:  
\begin{itemize}
	\item RF, SVR, and ensemble structure share most of the best results among themselves. There is no clearly superior method. 
	\item Bagging resampling performs slightly better than additive for short-term wind direction forecasting while for long-term ones, the reverse is true.
	\item For long-term forecasting, the ensemble structure mostly performs better than the alternatives. For instance, A-Sampling reaches $0.287$ MAAPE while ELM, as the best alternative, reaches $0.402$.        
\end{itemize}

\begin{table}
	\centering
	\caption{ MAAPE, NRMSE and MAE comparison for wind direction}
	\label{tab:forecastDir}
	\setlength\tabcolsep{3pt}
	\hspace*{-1cm}
	
	\begin{tabular}{llllllllllll}
		MAAPE &            &  &        &  &            &            &  &        &  &                     &               \\
		&            &  & ELM    &  & SVR        &            &  & RF     &  & Proposed Method     &               \\ \cline{4-4} \cline{6-7} \cline{9-9} \cline{11-12} 
		& Time Steps &  &        &  & $\epsilon$-SVR     & $\nu$-SVR     &  &        &  & A-Sampling (EF)         & B-Sampling (EF)   \\ \cline{2-12} 
		& 1          &  & 0.331  &  & 0.071      & 0.087      &  & 0.096  &  & 0.125 (1)           & 0.079 (6)     \\
		& 6          &  & 0.341  &  & 0.148      & 0.165      &  & 0.186  &  & 0.164 (1)           & 0.159 (6)     \\
		& 12         &  & 0.357  &  & 0.401      & 0.214      &  & 0.235  &  & 0.209 (1)           & 0.208 (2)     \\
		& 24         &  & 0.369  &  & 0.447      & 0.267      &  & 0.281  &  & 0.300 (1)           & 0.291 (2)     \\
		& 36         &  & 0.382  &  & 0.465      & 0.333      &  & 0.353  &  & 0.329 (2)           & 0.336 (2)     \\
		& 48         &  & 0.384  &  & 0.461      & 0.389      &  & 0.430  &  & 0.326 (2)           & 0.339 (6)     \\
		& 72         &  & 0.402  &  & 0.459      & 0.457      &  & 0.453  &  & 0.287 (2)           & 0.326 (1)     \\ \cline{2-2} \cline{4-12} 
		&            &  &        &  &            &            &  &        &  &                     &               \\
		NRMSE &            &  &        &  &            &            &  &        &  &                     &               \\
		&            &  & ELM    &  & \multicolumn{2}{l}{SVR} &  & RF     &  & \multicolumn{2}{l}{Proposed Method} \\ \cline{4-4} \cline{6-7} \cline{9-9} \cline{11-12} 
		& Time Steps &  &        &  & $\epsilon$-SVR     & $\nu$-SVR     &  &        &  & A-Sampling          & B-Sampling    \\ \cline{2-12} 
		& 1          &  & 20.407  &  & 7.252      & 7.928      &  & 8.752  &  & 11.653               & 7.772         \\
		& 6          &  & 21.669 &  & 14.554     & 15.797    &  & 15.194 &  & 14.884              & 14.430        \\
		& 12         &  & 22.141 &  & 19.369     & 21.382    &  & 18.730 &  & 18.228              & 17.911        \\
		& 24         &  & 23.603 &  & 22.009     & 23.958     &  & 23.902 &  & 21.486              & 20.963        \\
		& 36         &  & 43.000 &  & 24.412     & 27.882     &  & 26.365 &  & 29.516              & 28.198        \\
		& 48         &  & 23.073 &  & 27.570     & 31.508     &  & 25.503 &  & 28.820              & 29.223        \\
		& 72         &  & 92.813 &  & 28.714     & 28.061     &  & 25.904 &  & 30.861              & 28.464        \\ \cline{2-12} 
		&            &  &        &  &            &            &  &        &  &                     &               \\
		MAE   &            &  &        &  &            &            &  &        &  &                     &               \\
		&            &  & ELM    &  & \multicolumn{2}{l}{SVR} &  & RF     &  & \multicolumn{2}{l}{Proposed Method} \\ \cline{4-4} \cline{6-7} \cline{9-9} \cline{11-12} 
		& Time Steps &  &        &  & $\epsilon$-SVR     & $\nu$-SVR     &  &        &  & A-Sampling          & B-Sampling    \\ \cline{2-2} \cline{4-7} \cline{9-12} 
		& 1          &  & 62.555  &  & 11.842      & 15.265      &  & 16.343  &  & 27.414               & 13.457         \\
		& 6          &  & 64.623  &  & 27.597      & 32.112      &  & 34.049  &  & 32.926               & 29.093         \\
		& 12         &  & 68.037  &  & 39.502      & 47.211      &  & 46.758  &  & 41.046               & 40.378         \\
		& 24         &  & 71.275  &  & 50.738      & 55.979      &  & 65.802  &  & 55.057               & 54.217         \\
		& 36         &  & 78.023  &  & 62.537      & 71.256      &  & 73.336  &  & 78.593               & 79.029         \\
		& 48         &  & 72.834 &  & 74.934      & 86.684      &  & 72.169  &  & 76.020               & 80.233         \\
		& 72         &  & 90.937  &  & 85.218      & 80.023      &  & 75.403  &  & 70.961               & 76.431         \\ \cline{2-12} 
	\end{tabular}
\end{table}

One of the fundamental features of wind energy is the actual output power of the wind absorbed by the turbines. To investigate the performance of the methods for forecasting this feature, we use the same open-source dataset of GECAD that we used to conduct the seasonality experiment but this time we consider the first ten days of 2011 (or exactly $6 \times 24 \times 10 = 1440$ data points). Table \ref{tab:forecastPower} reports the power estimation results. One can observe:
\begin{itemize}
	\item The proposed method, considering both resampling techniques, clearly has an advantage over the others. Like previous tables, the deeper the prediction horizon is, greater the margin becomes.
	\item Most of the reported errors for 48 and 72 steps into the future are pretty close to the ones with shallower horizons. This may be due to the fact that all three error measurements have the sense of normalization inside them, which makes them comparable in separate contents, but we argue that it can be due to the repetitive patterns of the wind energy as well.
	\item The bootstrap aggregation technique performs best in short to medium horizons, while additive techniques have the best performance for long-term predictions. This was the case in the wind direction estimation, as well, with less intensity.         
\end{itemize}  

\begin{table}
	\centering
	\caption{ MAAPE, NRMSE and MAE comparison for wind power on GECAD data}
	\label{tab:forecastPower}
	\setlength\tabcolsep{3pt}
	\hspace*{-1cm}
	
	\begin{tabular}{llllllllllll}
		MAAPE &            &  &        &  &            &            &  &        &  &                     &               \\
		&            &  & ELM    &  & SVR        &            &  & RF     &  & Proposed Method     &               \\ \cline{4-4} \cline{6-7} \cline{9-9} \cline{11-12} 
		& Time Steps &  &        &  & $\epsilon$-SVR     & $\nu$-SVR     &  &        &  & A-Sampling (EF)         & B-Sampling (EF)   \\ \cline{2-12} 
		& 1          &  & 0.580  &  & 0.324      & 0.344      &  & 0.357  &  & 0.390 (1)           & 0.396 (2)     \\
		& 6          &  & 0.659  &  & 0.518      & 0.512      &  & 0.568  &  & 0.461 (2)           & 0.453 (1)     \\
		& 12         &  & 0.638  &  & 0.585      & 0.583      &  & 0.611  &  & 0.465 (1)           & 0.459 (1)     \\
		& 24         &  & 0.625  &  & 0.616      & 0.621      &  & 0.613  &  & 0.428 (2)           & 0.463 (1)     \\
		& 36         &  & 0.677  &  & 0.644      & 0.626      &  & 0.670  &  & 0.411 (2)           & 0.459 (2)     \\
		& 48         &  & 0.641  &  & 0.616      & 0.583      &  & 0.631  &  & 0.397 (2)           & 0.396 (1)     \\
		& 72         &  & 0.649  &  & 0.668      & 0.667      &  & 0.617  &  & 0.402 (2)           & 0.400 (2)     \\ \cline{2-12}  
		&            &  &        &  &            &            &  &        &  &                     &               \\
		NRMSE &            &  &        &  &            &            &  &        &  &                     &               \\
		&            &  & ELM    &  & \multicolumn{2}{l}{SVR} &  & RF     &  & \multicolumn{2}{l}{Proposed Method} \\ \cline{4-4} \cline{6-7} \cline{9-9} \cline{11-12} 
		& Time Steps &  &        &  & $\epsilon$-SVR     & $\nu$-SVR     &  &        &  & A-Sampling          & B-Sampling    \\ \cline{2-12} 
		& 1          &  & 23.619  &  &10.866     & 11.043      &  & 12.197 &  & 19.820              & 21.730        \\
		& 6          &  & 28.484 &  & 19.509     & 19.287      &  & 23.237 &  & 24.231              & 23.470        \\
		& 12         &  & 29.429 &  & 23.788     & 23.420      &  & 26.708 &  & 24.915              & 24.905        \\
		& 24         &  & 29.163 &  & 26.632     & 26.529      &  & 28.258 &  & 25.350              & 25.699        \\
		& 36         &  & 185.49 &  & 27.019     & 26.012      &  & 32.086 &  & 24.871              & 24.955        \\
		& 48         &  & 33.912 &  & 24.697     & 23.542      &  & 27.280 &  & 23.206              & 23.157        \\
		& 72         &  & 31.984 &  & 26.971     & 26.698      &  & 27.525 &  & 23.443              & 23.434        \\ \cline{2-12} 
		&            &  &        &  &            &            &  &        &  &                     &               \\
		MAE   &            &  &        &  &            &            &  &        &  &                     &               \\
		&            &  & ELM    &  & \multicolumn{2}{l}{SVR} &  & RF     &  & \multicolumn{2}{l}{Proposed Method} \\ \cline{4-4} \cline{6-7} \cline{9-9} \cline{11-12} 
		& Time Steps &  &        &  & $\epsilon$-SVR     & $\nu$-SVR     &  &        &  & A-Sampling          & B-Sampling    \\ \cline{2-2} \cline{4-7} \cline{9-12} 
		& 1          &  & 74.781  &  &  32.873      & 33.821      &  & 36.987  &  & 54.072               & 58.323         \\
		& 6          &  & 97.460  &  &  64.508      & 63.654      &  & 76.970  &  & 69.800               & 67.899         \\
		& 12         &  & 98.999  &  &  81.504      & 803.29      &  & 90.102  &  & 72.363               & 71.928         \\
		& 24         &  & 96.353  &  &  92.036      & 92.307      &  & 94.770  &  & 73.867               & 74.637         \\
		& 36         &  & 167.68  &  &  93.295      & 89.458      &  & 107.01  &  & 70.198               & 70.379         \\
		& 48         &  & 97.394  &  &  83.126      & 77.641      &  & 90.284  &  & 64.730               & 64.619         \\
		& 72         &  & 99.125  &  &  92.365      & 91.649      &  & 88.576  &  & 64.921               & 64.900         \\ \cline{2-12} 
	\end{tabular}
\end{table}

Furthermore, by considering all three tables together: \label{MAAPE true method}
\begin{itemize}
	\item Wind power prediction is the most challenging one and requires more diligence. Some of the methods' errors eventually become pretty enormous for power forecasting. Wind direction prediction stands second.
	\item MAAPE can be seen as a true unitless method across the separate tables. In addition, its value changes with viscosity and this makes its overall interpretation easier.
	\item The Extreme Learning machine performs worse compared to the other methods. Random forest is less consistent and experiences more fluctuations. It reaches the best performance in cases such as those of the NRMSE errors for wind direction, but performs the worst for wind speed regarding the same error measurement.
	\item SVR performs solid for short-term predictions but its performance fades away when considering deeper horizons.
	\item We suggest bagging resampling for short horizons and additive sampling for longer terms.        
\end{itemize}

Finally, we investigate the importance of the wind speed for forecasting the wind output power. In other words, we aim to explore the important question of how much the output power of wind renewable energy is based on historical data on the wind speed at that location. Hence, we only use wind power as the labels and not the input to the model. We still use the same dataset as the one for power prediction. Table \ref{tab:forecastWindToPower} indicates the comparisons. We can observe:
\begin{itemize}
	\item The errors of forecasting the wind output power based on wind speed are very close to the ones based on power historical data. This confirms the strong correlation between the two. The errors are even smaller for short horizons. Therefore, we suggest using wind speed to estimate accurate wind power along with historical power data. 
	\item ELM provides satisfactory performance for short-term cases, but it fails to maintain its superiority.
	\item RF and our proposed methods accomplish the best performance specifically when the horizon increases.
\end{itemize}

\begin{table}
	\centering
	\caption{ MAAPE, NRMSE and MAE comparison for wind power based on wind speed on GECAD data}
	\label{tab:forecastWindToPower}
	\setlength\tabcolsep{3pt}
	\hspace*{-1cm}

	\begin{tabular}{llllllllllll}
		MAAPE &            &  &        &  &            &            &  &        &  &                     &               \\
		&            &  & ELM    &  & SVR        &            &  & RF     &  & Proposed Method     &               \\ \cline{4-4} \cline{6-7} \cline{9-9} \cline{11-12} 
		& Time Steps &  &        &  & $\epsilon$-SVR     & $\nu$-SVR     &  &        &  & A-Sampling (EF)         & B-Sampling (EF)   \\ \cline{2-12} 
		& 1          &  & 0.276  &  & 0.332      & 0.325      &  & 0.369  &  & 0.384 (1)           & 0.387 (1)     \\
		& 6          &  & 0.569  &  & 0.511      & 0.516      &  & 0.553  &  & 0.488 (1)           & 0.471 (1)     \\
		& 12         &  & 0.570  &  & 0.569      & 0.577      &  & 0.604  &  & 0.468 (1)           & 0.468 (4)     \\
		& 24         &  & 0.624  &  & 0.617      & 0.617      &  & 0.603  &  & 0.443 (2)           & 0.463 (6)     \\
		& 36         &  & 0.634  &  & 0.631      & 0.628      &  & 0.658  &  & 0.402 (2)           & 0.459 (2)     \\
		& 48         &  & 0.589  &  & 0.592      & 0.585      &  & 0.610  &  & 0.381 (2)           & 0.454 (1)     \\
		& 72         &  & 0.635  &  & 0.674      & 0.659      &  & 0.630  &  & 0.398 (2)           & 0.458 (2)     \\ \cline{2-12}  
		&            &  &        &  &            &            &  &        &  &                     &               \\
		NRMSE &            &  &        &  &            &            &  &        &  &                     &               \\
		&            &  & ELM    &  & \multicolumn{2}{l}{SVR} &  & RF     &  & \multicolumn{2}{l}{Proposed Method} \\ \cline{4-4} \cline{6-7} \cline{9-9} \cline{11-12} 
		& Time Steps &  &        &  & $\epsilon$-SVR     & $\nu$-SVR     &  &        &  & A-Sampling          & B-Sampling    \\ \cline{2-12} 
		& 1          &  & 8.730  &  &10.778     & 10.761      &  & 12.248  &  & 16.285              & 20.219        \\
		& 6          &  & 21.639 &  & 19.341     & 19.429      &  & 22.293 &  & 25.231              & 22.870        \\
		& 12         &  & 23.838 &  & 22.797     & 23.185      &  & 25.166 &  & 25.140              & 25.116        \\
		& 24         &  & 27.409 &  & 26.547     & 26.312      &  & 28.092 &  & 26.474              & 25.708        \\
		& 36         &  & 28.656 &  & 26.393     & 26.237      &  & 30.687 &  & 25.621              & 24.885        \\
		& 48         &  & 24.959 &  & 23.716     & 23.472      &  & 26.683 &  & 23.916              & 23.156        \\
		& 72         &  & 26.263 &  & 27.262     & 26.331      &  & 26.865 &  & 25.743              & 23.449        \\ \cline{2-12} 
		&            &  &        &  &            &            &  &        &  &                     &               \\
		MAE   &            &  &        &  &            &            &  &        &  &                     &               \\
		&            &  & ELM    &  & \multicolumn{2}{l}{SVR} &  & RF     &  & \multicolumn{2}{l}{Proposed Method} \\ \cline{4-4} \cline{6-7} \cline{9-9} \cline{11-12} 
		& Time Steps &  &        &  & $\epsilon$-SVR     & $\nu$-SVR     &  &        &  & A-Sampling          & B-Sampling    \\ \cline{2-2} \cline{4-7} \cline{9-12} 
		& 1          &  & 26.959  &  &  32.803      & 32.500      &  & 38.394  &  & 46.896               & 54.200         \\
		& 6          &  & 74.355  &  &  63.622      & 64.163      &  & 73.477  &  & 73.863               & 67.364         \\
		& 12         &  & 79.267  &  &  77.863      & 79.386      &  & 86.233  &  & 73.045               & 72.989         \\
		& 24         &  & 93.151  &  &  91.947      & 91.365      &  & 93.178  &  & 77.506               & 74.657         \\
		& 36         &  & 94.274  &  &  90.645      & 90.001      &  & 102.60  &  & 71.839               & 70.230         \\
		& 48         &  & 81.849  &  &  78.804      & 77.739      &  & 86.790  &  & 65.302               & 64.615         \\
		& 72         &  & 88.215  &  &  93.654      & 90.036      &  & 88.677  &  & 69.010               & 64.935         \\ \cline{2-12} 
	\end{tabular}
\end{table}

\subsubsection{Feature Selection}
\label{sec:feat}
In this section we aim to tackle the concept of feature selection. We develop a framework to select the important inputs affecting wind power and throw away unsatisfying ones. We bring SDFO-TR one more time into the design of the framework, and not only for tuning. For each feature, we introduce a variable and we let SDFO-TR decide whether to use the feature's data. This injects all the complexity into the DFO optimization. Therefore, the DFO performance can be affected when it encounters a large number of features. However, the process benefits from clear interpretation. In other words, SDFO-TR ideally identifies the exact list of dominant features. It is worth mentioning that the whole feature selection process is automated. In the algorithm \ref{alg:1}, we increase the number of variables as $x =(x^1,...,x^{(P+Q)}),$ where $Q$ is the total number of features and $x^i \in \left[0,1\right], \ \forall i=P+1,..., P+Q,$ with $\dfrac{1}{2}$ being considered as the threshold for whether to choose the feature. Figure \ref{framework} illustrates the proposed method in a general view. The SDFO-TR algorithm chooses the features and parameters for the ensemble structure up to the budget $\mathcal{L}$.

We want to investigate the effect of different environmental characteristics on wind speed and therefore indirectly on wind power. This is not a prediction since we do not use wind speed historical data as an input to the process, but only as the labels to evaluate the errors. That is, we are certain about the fact that historical data on wind speed can greatly enhance its prediction, and Table \ref{tab:forecastSpeed} is in fact dedicated to this. However, to correctly investigate other major factors, we cast aside wind speed from the inputs and only use it as the output label for accuracy evaluations.

To this end, we focus on another open-source GECAD dataset {\footnote{From the site: \href{http://sites.ieee.org/pes-iss/data-sets/}{Power \& Energy Society Open Data Sets (http://sites.ieee.org/pes-iss/data-sets/)}; under the Weather Data section; gecad-weather; Porto, Portugal, 2016.}}, which is rich in the number of presented features. The features are [Temperature (C), Dew-point (C), Pressure (Pa), Wind Direction, Wind Direction Degrees ($\degree$), Wind Speed (KMH), Wind Gust (KMH), Humidity, Hourly Precipitation (MM), daily rain (MM), Solar Radiation (Watt/$m^2$)]. We always choose Wind Direction Degrees column so that we have at least one feature. We further omit the column Wind Direction because it is a duplicate of Wind Direction Degrees column but expressed qualitatively. We focus on the first ten days of the dataset due to computational limitations and the further train/test division routines are as same as previous experiments. All the models have access to up-to-date measurements at each time step except wind speed which is under study. To the best of the authors' knowledge, this comparison to find the most effective elements on wind speed has not been studied.

Table \ref{tab:FS1} represents the effect of using SDFO-TR for feature selection as well as a comparison with Spearmint Bayesian optimization hyperparameter tuning method. We compare the presented method with and without feature selection for both of the resampling techniques. The Spearmint is designed to tune parameters of one sequence-to-sequence network. We tune all the parameters considered in the proposed method with Spearmint except the EF. We internationally focus on one sequence-to-sequence network to see the impact of EF on the performance of the proposed structure. In other words, the results showing in Table \ref{tab:FS1}, indicate the differences between DFO and Spearmint hyperparameter methods combined with the effect of using ensemble of sequence-to-sequence networks versus single network.

Throughout the paper, we report three different error metrics. One can consider any of these metrics as the loss function of DFO or Spearmint. So far in all the studies, we considered MAAPE as the main objective function of DFO and then we obtain the predictions. After we obtained the predictions and the optimization process is done, we calculate the other two errors based on the predictions for report purposes. The three errors are mostly in a harmony having DFO as the tuning method. However, We realize it is not the case for Spearmint method. In other words, we may optimize the process considering MAAPE and reach a satisfying MAAPE error but at the same time we obtain relatively large value for NRMSE or MAE. Having said this, we conduct three experiments with Spearmint for each of the three error metrics as its objective function.

From the information of Table \ref{tab:FS1} one can argue:
\begin{itemize}
	\item Methods with the selected group of features have superior performance compared to those using all the information.
	\item For ensemble structure, additive resampling in general is more robust compare to bagging.
	\item Spearmint performance is promising when we consider MAE and MAAPE as its objective function but its performance is inferior compare to other methods when we use NRMSE as its objective function. Moreover, when we optimize Spearmint with MAAPE, it reaches the best accuracy between all the methods but simultaneously it reports not a satisfying NRMSE. We conclude that Spearmint exploits the tuning process with one objective function to such an extent that it can generally result in inferior values for other objectives.
	\item The most robust performance of Spearmint is when we optimize MAE as its objective. In this case, the results are in a close tie with the proposed structure with Additive resampling.      
	\item Temperature, Dew-point, wind gust and solar radiation are mostly chosen by different methods which shows their important correlations with wind speed. It is natural to guess that wind gust contains information about wind speed. Please note that we always select wind wind direction degrees column.  
	\item Pressure is mainly not chosen in the selected group by majority of the methods. Therefore, we suggest there is less correlation between wind speed and pressure.      
\end{itemize}  

\begin{table}
	\centering
	\caption{ Feature selection for wind speed on GECAD data}
		\vspace{1em}
	\label{tab:FS1}
	\begin{tabular}{lllllllllll}
		Method                       &  & EF   &  & MAAPE          &  & NRMSE           &  & MAE            &  & Selected Features   \\ \cline{1-1} \cline{3-3} \cline{5-5} \cline{7-7} \cline{9-9} \cline{11-11} 
		Additive (all feature)       &  & 1    &  & 0.834          &  & 11.025          &  & 2.569                   &  & [1,1,1,1,1,1,1,1,1] \\                
		Bagging (all feature)        &  & 3    &  & 0.943          &  & 11.028          &  & 3.615                   &  & [1,1,1,1,1,1,1,1,1] \\
		Additive (feature Selection) &  & 5    &  & 0.826          &  & 11.013          &  & 2.490                   &  & [1,1,0,1,1,0,1,1,1] \\
		Bagging (feature Selection)  &  & 2    &  & 0.924          &  & 9.751           &  & 3.199                   &  & [1,1,0,1,1,0,0,0,1] \\ 
		Spearmint (Obj. function: MAAPE)      &  & 1    &  & 0.690          &  & 12.225          &  & 3.152          &  & [1,1,0,1,1,0,0,0,1] \\ 
		Spearmint (Obj. function: NRMSE)      &  & 1    &  & 0.805          &  & 17.337          &  & 4.778          &  & [1,0,1,1,1,1,0,1,1] \\		
		Spearmint (Obj. function: MAE)        &  & 1    &  & 0.846          &  & 11.104          &  & 2.441          &  & [0,1,0,1,1,1,0,0,0] \\				
\end{tabular}
\end{table}

\subsubsection{Online Forecasting}
\label{sec:online}

So far, all of the studies we have discussed are conducted in an offline setting which means we train the model on a train dataset and then test its performance on the test set. On the other hand, in online settings the model simultaneously learns through the dataset with information being revealed one step at a time. Here, we briefly address the online forecasting setting. Any online method estimates (mostly) one step ahead into the future, and then the realization of that forecast happens subsequently and the model has access to it. Afterwards, the model estimates the next point and the online process goes on. To this end, the division of the data into train, validation and test does not make sense. Instead, the model updates its learning during the process at each time step. In other words, the model learns from some initial data and then learns from and is evaluated by the rest of the data simultaneously. Having said this important difference, we still call the second part of the dataset as the test set, but note that the model learns through the test set as well.

In this setting, one can modify the two resampling techniques mentioned above, but we simplify our model to have only one sequence-to-sequence network. Therefore, there is no resampling used in this section. Once again we utilize the GECAD dataset exactly as we used it for the prediction of wind power output in Table \ref{tab:forecastPower}. This time, the entire test set, or equivalently the last 20 percent of the whole dataset, is considered as the online setting test set. We do this so that we can compare the errors in regular forecasting with online case. 

Table \ref{tab:OF} reports the comparisons between the proposed method and the SARIMA and FOSELM methods that were discussed in Section \ref{Alt-Alg}. It is worth mentioning that for every single point in the online test set, we fit a whole new SARIMA model and estimate one step ahead into the future, and at the end, we evaluate the errors. We use a buffer of 100 points for each of these SARIMA models, and disregard the rest of the dataset because more than this buffer size only complicates the SARIMA structure and does not help with the performance. We found that FOSELM method very sensitive to the tuning of its parameters. Therefore, the reported results for this method are the best ones out of 10 independent runs. We can observe:
\begin{itemize}
	\item The proposed Method outperforms the other two by a significant margin.
	\item The proposed Method results are better compared to the similar case in regular forecasting in Table \ref{tab:forecastPower}. The reason is that in the online settings, we allow the model to learn during the testing period.
\end{itemize}      

\begin{table}
	\centering
	\caption{Online Forecasting for wind power}
	\vspace{1em}
	\label{tab:OF}
	\begin{tabular}{lllllll}
		Method          &  & MAAPE          &  & NRMSE           &  & MAE             \\ \cline{1-1} \cline{3-3} \cline{5-5} \cline{7-7} 
		FOSELM          &  & 0.614          &  & 25.272          &  & 82.644          \\
		SARIMA          &  & 0.424          &  & 17.272          &  & 50.649          \\
		Proposed Method &  & \textbf{0.280} &  & \textbf{10.878} &  & \textbf{31.597} \\ \hline
	\end{tabular}
\end{table}

\begin{figure}
	\hspace*{-1.25cm}
	\scalebox{0.40}{\input{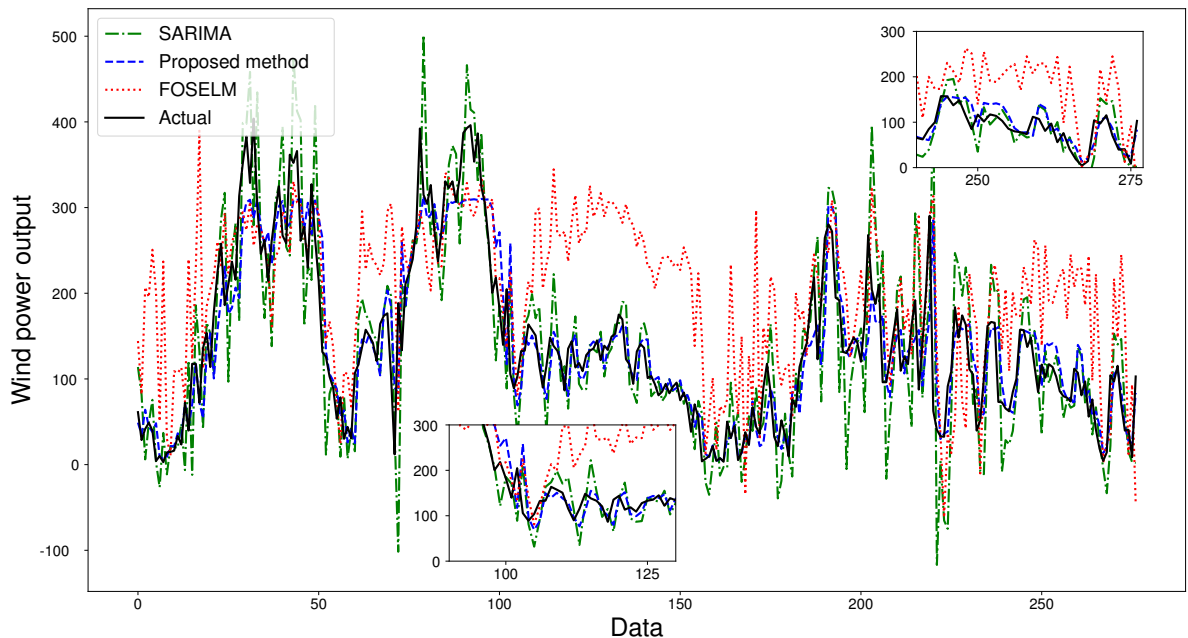}}
	\caption{Wind power online forecasting. Inset boxes illustrate zoomed-in portions}
	\label{fig:22}
\end{figure}

Figure \ref{fig:22} demonstrates actual test set along with the online forecasting results of the three methods. We mention one more time that after each prediction, the models observe the ground truth values of that time step and they use it in the next prediction. We can see that SARIMA and our method have a clear advantage over FOSELM. In addition, both SARIMA and FOSELM estimate some negative values for the output power which indicate the considerable gap between the actual non-negative output power and the predictions from these two methods.   

\subsection{Solar Renewable Energy Forecasting}
Converted into electrical and thermal energy, solar renewable energy is one of the most promising alternatives for fossil fuels, originating from radiant light and heat from the Sun. In terms of global capacity, solar Photovoltaics (PV) is among the first three renewable sources besides Hydroelectricity and wind powers \cite{kalogirou2013solar}. Here, we focus on prediction of global horizontal irradiance (GHI) which is the total amount of terrestrial radiation falling on a surface horizontal to the surface of the ground. GHI is of a certain importance for photovoltaic conversion processes because it is used for PV output power calculation and it accounts for both of the direct normal irradiance (DNI) and diffuse horizontal irradiance (DHI) \cite{ameen2019modelling}. To this end, we use an open source solar data provided by national solar radiation database (NSRDB) \cite{sengupta2018national} for central park New York located at $(40.77, -73.98)$ coordination for year 2017 \footnote{NSRDB data viewer can be found at this \href{https://maps.nrel.gov/nsrdb-viewer/?aL=UdPEX9\%255Bv\%255D\%3Dt\%26f69KzE\%255Bv\%255D\%3Dt\%26f69KzE\%255Bd\%255D\%3D1&bL=clight&cE=0&lR=0&mC=4.740675384778373\%2C22.8515625&zL=2}{site}.}. The dataset contains measurements thirty minutes apart each. The raw dataset roughly contains 50 percent zero GHI which accounts for the hours where the exact location does not receive any radiant light from the Sun. We first conduct some forecastings on the raw dataset without omitting any zeros. Then, we eliminated all the zeros and conduct the same experiment. Comparisons confirm that eliminating the zeros significantly improves the accuracy of the prediction. We consider the first 4320 data points (three first months of 2017). Considering MAAPE as the error metric, Table \ref{tabzeroNonZero} indicates the comparison between the case where we considered all the measurements fore forecasting versus the case where we eliminated the zeros. The datasets for the two experiences are not exactly the same but MAAPE as previously mentioned in Section \ref{MAAPE true method} is the most robust error metric among the metrics we studied and it is reliable for relative comparisons. As one can see in Table \ref{tabzeroNonZero}, the MAAPE errors reported for the reduced dataset are noticeably less than the raw dataset. Therefore, for the comprehensive comparisons with alternative methods in the following table, we focus on the reduced dataset. 

\begin{table}[]
		\caption{MAAPE error comparison between GHI Raw and reduced datasets.}
	\label{tabzeroNonZero}
	\vspace{1em}
\centering
\begin{tabular}{ccccccccc}
	\hline
	&  & \multicolumn{3}{c}{\textbf{Additive Resampling}} &  & \multicolumn{3}{c}{\textbf{Bootstrap Resampling}} \\
	&  & \multicolumn{3}{c}{MAAPE Error}                  &  & \multicolumn{3}{c}{MAAPE Error}                   \\ \cline{3-5} \cline{7-9} 
	Time Steps &  & Raw Dataset       &       & Reduced Dataset      &  & Raw Dataset       &        & Reduced Dataset      \\ \cline{1-1} \cline{3-3} \cline{5-5} \cline{7-7} \cline{9-9} 
	1          &  & 1.020             &       & 0.425                &  & 1.112             &        & 0.500                \\
	6          &  & 1.025             &       & 0.605                &  & 1.167             &        & 0.624                \\
	12         &  & 1.114             &       & 0.627                &  & 1.192             &        & 0.621                \\
	24         &  & 1.174             &       & 0.615                &  & 1.206             &        & 0.624                \\
	36         &  & 1.158             &       & 0.620                &  & 1.194             &        & 0.626                \\
	48         &  & 1.121             &       & 0.594                &  & 1.190             &        & 0.624                \\
	72         &  & 1.209             &       & 0.623                &  & 1.216             &        & 0.623                \\ \hline
\end{tabular}
\end{table}
We explore GHI estimation and compare the results of the proposed method with SVR and RF. Table \ref{tab:forecastGHI} reports the findings. 
As shown previously, the proposed method considering both resampling techniques is more robust against increasing the forecasting horizon. SVR and RF methods have advantages over the method for the very short forecasting steps, but for longer forecasting horizons, the method performs relatively better. Moreover, the reported MAAPE error for GHI even on reduced dataset is higher than reported MAAPE error of wind direction and speed and is relatively close to errors mentioned for forecasting wind power. That is, the stochasticity of the GHI is considerably high.

\begin{table}
	\centering
	\caption{ MAAPE, NRMSE and MAE comparison for global horizontal irradiance (GHI)}
		\vspace{1em}
	\label{tab:forecastGHI}
	\setlength\tabcolsep{3pt}
	\hspace*{-1cm}
	\begin{tabular}{llllllllllll}
		MAAPE &            &  &        &  &            &            &  &        &  &                     &               \\
		&                &  & SVR        &            &  & RF     &  & Proposed Method     &               \\ \cline{4-5} \cline{7-7} \cline{9-12}  
		& Time Steps         &  & $\epsilon$-SVR     & $\nu$-SVR     &  &        &  & A-Sampling (EF)         & B-Sampling (EF)   \\ \cline{2-12} 
		& 1            &  & 0.358      & 0.359      &  & 0.345  &  & 0.425 (1)           & 0.500 (4)     \\
		& 6            &  & 0.634      & 0.640      &  & 0.603  &  & 0.605 (1)           & 0.624 (1)     \\
		& 12           &  & 0.676      & 0.724      &  & 0.687  &  & 0.627 (1)           & 0.621 (1)     \\
		& 24           &  & 0.697      & 0.718      &  & 0.692  &  & 0.615 (3)           & 0.624 (1)     \\
		& 36           &  & 0.739      & 0.743      &  & 0.728  &  & 0.620 (3)           & 0.626 (4)     \\
		& 48           &  & 0.717      & 0.730      &  & 0.717  &  & 0.594 (6)           & 0.624 (4)     \\
		& 72           &  & 0.726      & 0.768      &  & 0.736  &  & 0.623 (1)           & 0.623 (2)     \\ \cline{2-12}  
		&                    &  &            &            &  &        &  &                     &               \\
		NRMSE &            &  &        &  &            &            &  &        &  &                     &               \\
		&             &  & \multicolumn{2}{l}{SVR} &  & RF     &  & \multicolumn{2}{l}{Proposed Method} \\ \cline{4-5} \cline{7-7} \cline{9-12}
		& Time Steps  &  & $\epsilon$-SVR     & $\nu$-SVR     &  &        &  & A-Sampling          & B-Sampling    \\ \cline{2-12} 
		& 1           &  &10.855     & 10.624      &  & 11.731  & & 24.226              & 35.504        \\
		& 6           &  &27.105     & 26.500      &  & 23.963 &  & 34.576              & 39.674        \\
		& 12          &  &30.675     & 37.012      &  & 28.449 &  & 39.795              & 39.145        \\
		& 24          &  &35.442     & 38.247      &  & 31.311 &  & 40.790              & 40.122        \\
		& 36          &  &37.697     & 37.738      &  & 34.518 &  & 40.976              & 40.312        \\
		& 48          &  &37.245     & 36.398      &  & 35.113 &  & 38.064              & 39.961        \\
		& 72          &  &34.407     & 36.813      &  & 37.015 &  & 39.221              & 39.157        \\ \cline{2-12} 
		&             &  &            &            &  &        &  &                     &               \\
		MAE   &                 &            &  &        &  &                     &               \\
		&            &  & \multicolumn{2}{l}{SVR} &  & RF     &  & \multicolumn{2}{l}{Proposed Method} \\ \cline{4-5} \cline{7-7} \cline{9-12}
		& Time Steps &  & $\epsilon$-SVR     & $\nu$-SVR     &  &        &  & A-Sampling          & B-Sampling    \\ \cline{2-2} \cline{4-7} \cline{9-12} 
		& 1           &  &  66.248      & 63.7105      &  & 74.513  &  & 145.236               & 213.343         \\
		& 6           &  &  191.277      & 188.178      &  & 169.543  &  & 222.690               & 249.781         \\
		& 12          &  &  215.607      & 255.722      &  & 207.938  &  & 251.782               & 247.761         \\
		& 24          &  &  242.326      & 261.320      &  & 224.981  &  & 259.683               & 253.450         \\
		& 36          &  &  269.622      & 270.761      &  & 247.277  &  & 256.896               & 253.009         \\
		& 48          &  &  254.923      & 255.712      &  & 246.723  &  & 229.296               & 249.469         \\
		& 72          &  &  240.599      & 263.872      &  & 255.920  &  & 240.421               & 240.045         \\ \cline{2-12} 
	\end{tabular}
\end{table}

\subsection{Feature Selection in Wave Renewable Energy}
As long as wind blows with sufficient stability to provide continuous ocean waves, marine power contains promising potential as alternative energy \cite{lehmann2017ocean}. The marine infrastructure for harnessing such a power, however, is less developed compare to wind and solar sources at the time \cite{aderinto2018ocean}. One of the main reasons is that there are few studies investigating major contributors to ocean wave output power. Therefore, in this section we aim to find the most important marine features affecting the wave output power. We use the case study established by \cite{pirhooshyaran2019multivariate} in the east coast of the U.S.\footnote{ The data is obtained from national oceanic and atmospheric administration (\href{https://www.noaa.gov/}{NOAA}) and is available at \href{https://github.com/mamadpierre/NOAA-Refined-Stations}{this} GitHub repository.} The case study explores the effect of different ocean features, recorded by several sensors of different distances, on the significant wave height and ocean power of specified location. They use an Elastic net (EN) regularizer for feature selection. Similar to the study, we consider 53 features of 5 different nearby buoys. For more details on the buoys and their selected features, please see the Appendix \ref{appendix_B}. As far as the authors' knowledge extend, this study is the first study to utilize DFO as a designer in the concept of ocean wave energy.

To demonstrate a fair comparison of DFO method with other tuning/feature selection methods, we consider the exact ensemble structure of the networks as well as the exact marine features for a case where Spearmint is responsible for all the designing and tuning tasks. First, we consider 53 binary variables indicate whether we use any combination of the features. Second, we reserve an integer variable for EF and finally, we take into consideration the structural features of the sequence-to-sequence network such as number of stacked LSTM layers, learning rate and etc. The total number of features ends up to be 62 for both DFO and Spearmint. The data size is 3422 points. Therefore, the feature selection is conducted on a dataset of $3422 \times 62$.
We refer the reader to Appendix \ref{appendix_B}, to see the selected group of marine features for both DFO and Spearmint techniques. Table \ref{caseStudy53} indicates the results of the comparison. First two rows are reported from \cite{pirhooshyaran2019multivariate}. Furthermore, to be consistent with the mentioned study we used RMSE and HUBER losses besides MAAPE. In addition, we limit the experiments to additive resampling for DFO and Spearmint.      

\begin{table}[]
			\caption{Ocean wave power output.}
	\label{caseStudy53}
	\centering
		\vspace{1em}
	\begin{tabular}{ccccccc}
		&  & \multicolumn{5}{c}{Wave Power Output} \\ \cline{3-7} 
		\textbf{Technique} &  & MAAPE         &    & RMSE   &    & HUBER   \\ \cline{1-1} \cline{3-3} \cline{5-5} \cline{7-7} 
		RF                 &  & 0.56          &    & 3.43   &    & 5.62    \\ \cline{1-1}
		Elastic Net        &  & 0.45          &    & 4.53   &    & 2.21    \\ \cline{1-1}
		Spearmint          &  & 0.40          &    & 2.88   &    & 1.08        \\ \cline{1-1}
		DFO                &  & 0.35          &    & 3.15   &    & 1.23       \\ \hline
	\end{tabular}
\end{table}

We observe:
\begin{itemize}
	\item DFO and Spearmint reject using many features and this indicates the importance of feature selection for renewable energy concept. The most important features to foresee the ocean wave output power of an unknown location are its nearby air temperature, significant wave height and wind speed. (See Appendix \ref{appendix_B}).
	\item The two ensemble methods (DFO and Spearmint) have considerable advantage over the other two methods. The results suggest using of the parallel structures. 
	\item There are advantages and disadvantages comparing EN with DFO and Spearmint techniques. EN can partially utilize any input feature and in the cases where the datasets contain valuable information in many separate features, it can be helpful. DFO and Spearmint, however, accept or reject entire data related to a feature. Findings confirm that using the latter approach results in better accuracy. One can argue that stochasticity involved in the nature of the renewable sources can be one of the reasons that DFO and Spearmint omit many features. 
	\item In contrast to wind speed feature selection study, DFO rejects using of many of the features. This can be a confirmation on the fact that ocean waves are irregular waves with few solid correlations between them and other marine features.     
\end{itemize}

\section{Conclusion}
\label{sec:conclu}

We introduce an ensemble of sequence-to-sequence neural networks and combine the structure with derivative-free optimization and other tuning/designing techniques such as Spearmint Bayesian optimization to tackle both the forecasting and feature selection of any multivariate time series including but not limited to renewable energy sources. We introduce additive resampling technique to take into account the repetitive patterns of the data and to further utilize the parallel structure to its maximum extent possible by assigning different parts of the training set to separate networks. In addition to additive resampling, the well-established bagging resampling is considered throughout the paper.

The findings confirm the parallel structure can be successfully integrated with other optimization techniques such as DFO and Spearmint. Moreover, results indicate the superiority of the proposed method compare to many established machine learning approaches in particular in cases where one targets deeper steps into the unknown future.

Then, we present three independent case studies about wind, solar and wave renewable energy sources. All three sources are among the most affordable and scalable renewable sources but their ever-changing nature requires a framework to foresee its future outcomes. To this end, we forecast wind speed, wind direction, wind power output, global horizontal irradiance along with ocean wave power output. We realize wind output power and global horizontal irradiance have marginally bigger measurement errors compare to wind speed and direction which means more irregularity in wind power and global irradiance. In addition, we explore the effect of wind speed on forecasting wind output power. Results suggest that wind speed is actually a reliable indicator for the possible wind output power in particular for short term forecasts. This confirms a solid correlation between wind speed of a location and the produced wind output power. Moreover, we study forecasting of global horizontal irradiance. We suggest reducing the GHI datasets into nonzero measurements for better performance.

Then, we aim to find the most effective environmental and marine features on wind speed and ocean wave output power respectively. To this end, we modify the framework to automatically select major features and discard those that are not useful to estimation of the under study feature. Studies show temperature, Dew-point, wind gust and solar radiation are four most important features affecting wind speed which are chosen by majority of the methods. Significant wave height, wind speed and air temperature are among the most principal marine features affecting ocean wave output power. Moreover, we investigate online forecasting concept. The results substantiate the superiority of the method for forecasting wind power compare to other methods.

Ensemble network designs are promising structures and we suggest investigation of those in other aspects of renewable energy systems. Furthermore, employing DFO as a decision maker for the design of a parallel structure can be explored in many other machine learning disciplines.

\begin{appendices}
	
\section{LSTM and Sequence-to-Sequence Structures}\label{appendix_A}

We consider $[a_1,a_2,...,a_T]$ as an input sequence of an LSTM cell. For an LSTM cell for any time step $l=1,..,T$, we have \cite{greff2016lstm}:

\begin{alignat}{1}
\label{forget}
&	f_l = \text{sigm}   \left( W_{fa}a_l+ W_{fh}h_{l-1}  \right) \\ \label{input}
&	p_l = \text{sigm} \left(W_{pa}a_l+ W_{ph}h_{l-1} \right)\\
&	o_l = \text{sigm} \left(W_{oa}a_l+ W_{oh}h_{l-1} \right)\\ \label{state}
&	c_l = f_l \odot c_{l-1} + p_l \odot  \text{tanh} \left(W_{ca}a_l+ W_{ch}h_{l-1} \right)\\ \label{hidden}
&	h_l = o_l \odot \text{tanh} (c_l), 	  
\end{alignat}
where $\odot$ means element-wise matrix product, $ a_l \in \mathbb{R}^Q$, $ h_l \in \mathbb{R}^H$ are the input and hidden vectors, $ f_l \in \mathbb{R}^H$, $ p_l \in \mathbb{R}^H$, $ o_l \in \mathbb{R}^H$, $ c_l \in \mathbb{R}^H$ are forget, input, output and cell gates respectively, and at last $Q$ and $H$ are the input feature and LSTM hidden sizes. The actual output of each LSTM cell is its hidden layer.	

A sequence-to-sequence network contains two separate recurrent structures which in our case are two LSTMs. First structure (encoder) reads the input sequence $[a_1,a_2,...,a_T]$ considering $h_0=c_0=0$  and gives $ h_T$ vector as its output. Second structure (decoder) uses $ h_T$ as its initial state and deliver the prediction $[b_{T+T'}]$.

\section{Ocean Wave Case Study}\label{appendix_B}

\begin{figure}
	\centering
		\includegraphics[width = 12cm]{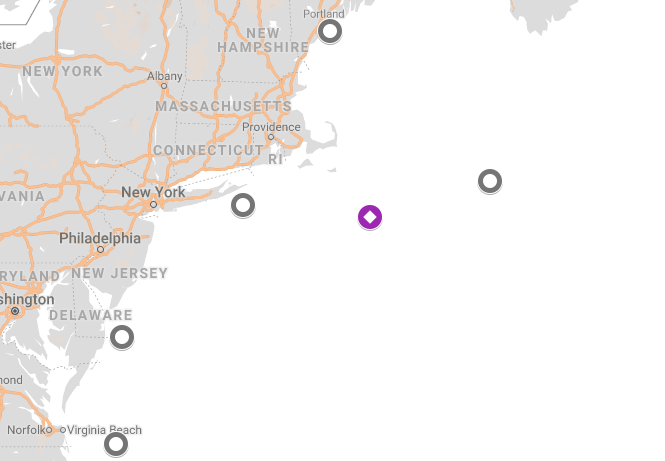}
	\caption{NOAA measurement stations for Ocean Wave power output Case study conducted by \cite{pirhooshyaran2019multivariate}.}
	\label{fig:caseStudy53}
\end{figure}

Figure \ref{fig:caseStudy53} schematically shows six active measurement stations of NOAA. Table \ref{tab1} tabulates the necessary information of the measurement buoys. This experiment is designed to see the effect of other ocean features on possible output power at a different location. We aim to forecast the ocean output power at station 44008 with the features from the other 5 adjacent buoys as mentioned in Table \ref{tab000}. The abbreviations of the features are explained in Table \ref{tab00}.

\begin{table}
	\caption{NOAA measurement stations}
	\label{tab1}
	\vspace{1em}
	\setlength\tabcolsep{3pt}
\centering
	\begin{tabular}{ccccccccc}
		\hline
		& \textbf{Station ID number} &  & 44007          & 44008          & 44009          & 44013          & 44014          & 44017          \\ \cline{2-2}
		& \textbf{Coordination}       &  & (70.14, 43.52) & (69.24, 40.50) & (74.70, 38.45) & (70.65, 42.34) & (74.84, 36.60) & (72.04, 40.69) \\ \cline{2-2}
		& \textbf{Depth of water}    &  & 26.5           & 74.7           & 30             & 64             & 47             & 48             \\ \hline
	\end{tabular}
	
\end{table}

\begin{table}[]\footnotesize
			\caption{53 input features and output label}
	\label{tab000}
	\setlength\tabcolsep{3pt}
	\hspace*{-1.25cm}
\begin{tabular}{lllllllllllllll}
			\hline
	Features &  & Station 44007: &  & WDIR7        & WSPD7  & GST7  & WVHT7  & DPD7  & APD7  & MWD7  & PRES7  & ATMP7  & WTMP7  & DEWP7  \\ \cline{1-1} \cline{3-3}
	         &  & Station 44009: &  & WDIR9        & WSPD9  & GST9  & WVHT9  & DPD9  & APD9  & MWD9  & PRES9  & ATMP9  & WTMP9  &        \\ \cline{3-3}
           	 &  & Station 44013: &  & WDIR13       & WSPD13 & GST13 & WVHT13 & DPD13 & APD13 & MWD13 & PRES13 & ATMP13 & WTMP13 &        \\ \cline{3-3}
	         &  & Station 44014: &  & WDIR14       & WSPD14 & GST14 & WVHT14 & DPD14 & APD14 & MWD14 & PRES14 & ATMP14 & WTMP14 & DEWP14 \\ \cline{3-3}
	         &  & Station 44017: &  & WDIR17       & WSPD17 & GST17 & WVHT17 & DPD17 & APD17 & MWD17 & PRES17 & ATMP17 & WTMP17 & DEWP17 \\ \cline{3-3}
	Label    &  & Station 44008: &  & Output power &        &       &        &       &       &       &        &        &        &        \\ \hline
\end{tabular}
\end{table}
\begin{table}[H]
	\centering
		\caption{Ocean wave features description}
	\label{tab00}
	\setlength\tabcolsep{3pt}
\centering
	\begin{tabular}{ccccccc}
		\hline
		\multicolumn{1}{c}{\textbf{Abbreviation}} &  & \textbf{Description}                    &  & \textbf{Abbreviation} &  & \multicolumn{1}{c}{\textbf{Description}}           \\ \cline{1-1} \cline{3-3} \cline{5-5} \cline{7-7} 
		\multicolumn{1}{c}{WDIR}                  &  & Wind direction                          &  & WSPD                  &  & \multicolumn{1}{c}{Wind speed (m/s)}               \\ \cline{1-1} \cline{3-3} \cline{5-5} \cline{7-7} 
		\multicolumn{1}{c}{WVHT}                  &  & Significant wave height (meters)        &  & DPD                   &  & \multicolumn{1}{c}{Dominant wave period (seconds)} \\ \cline{1-1} \cline{3-3} \cline{5-5} \cline{7-7} 
		\multicolumn{1}{c}{MWD}                   &  & Direction of DPD                        &  & PRES                  &  & \multicolumn{1}{c}{Sea level pressure (hPa)}       \\ \cline{1-1} \cline{3-3} \cline{5-5} \cline{7-7} 
		\multicolumn{1}{c}{GST}                   &  & GST Peak 5 or 8 second gust speed (m/s) &  & APD                   &  & \multicolumn{1}{c}{Average wave period (seconds)}  \\ \cline{1-1} \cline{3-3} \cline{5-5} \cline{7-7} 
		\multicolumn{1}{c}{WTMP}                  &  & Sea surface temperature (Celsius)       &  & DEWP                  &  & \multicolumn{1}{c}{Dewpoint temperature}           \\ \cline{1-1} \cline{3-3} \cline{5-5} \cline{7-7} 
		\multicolumn{1}{c}{ATMP}                  &  & Air temperature (Celsius)               &  &                       &  & \multicolumn{1}{c}{}                               \\ \hline
		&  &                                         &  &                       &  &                                                     \\
		&  &                                         &  &                       &  &                                                    
	\end{tabular}
\end{table}

The selected group of marine features for DFO is [WVHT7, WVHT9, WSPD13, WVHT13, WVHT14, WVHT17] and the rest are not selected. The selected group of features for Spearmint is
[WSPD7, WVHT7, APD7, PRES7, ATMP7, DEWP7, WSPD9, WVHT9, MWD9, PRES9, ATMP9, WDIR13, GST13, APD13, PRES13, ATMP13, WTMP13, WSPD14, WVHT14, DPD14, APD14, ATMP14, WDIR17, WSPD17, DPD17, ATMP17].
Therefore, air temperature, significant wave height and wind speed are the most gainful features to consider. \\
\end{appendices}

{\bf Acknowledgments}

This research was partially supported by the National Science Foundation via the CyberSEES grant \#1442858. The authors are accountable for the conclusions documented in the paper and the points claimed in the article have not been endorsed by the sponsoring agency.

\bibliography{ensemblePlusDFO}

\bibliographystyle{unsrtnat}

\end{document}

%% file: framework.tex
	
	\tikzstyle{decision} = [diamond,   draw, fill=white!20, text width=6em,  text centered,  node distance=2cm, inner sep=0pt]
	\tikzstyle{block}    = [rectangle, draw, fill=white!20, text width=9em,  text centered,                     rounded corners, minimum height=2em]
	\tikzstyle{blockPic} = [rectangle, draw, fill=white!20, text width=7em,  text centered,                     rounded corners, minimum height=2em]
	\tikzstyle{block1}   = [rectangle, draw, fill=white!20, text width=25em, text centered,                     rounded corners, minimum height=2em]
	\tikzstyle{line}     = [draw, -latex']

	\centering
	
	\begin{tikzpicture}[node distance = 1.5cm, auto]
	\node [block, ,outer sep=0, align=center] (Data) {Data};
	\node [block, align=center,below of= Data] (Resample) {Resampling};

	\node [matrix, below left=of Resample] (A) {\node [block] at (3,2) (additive) {  \includegraphics[width = 3cm]{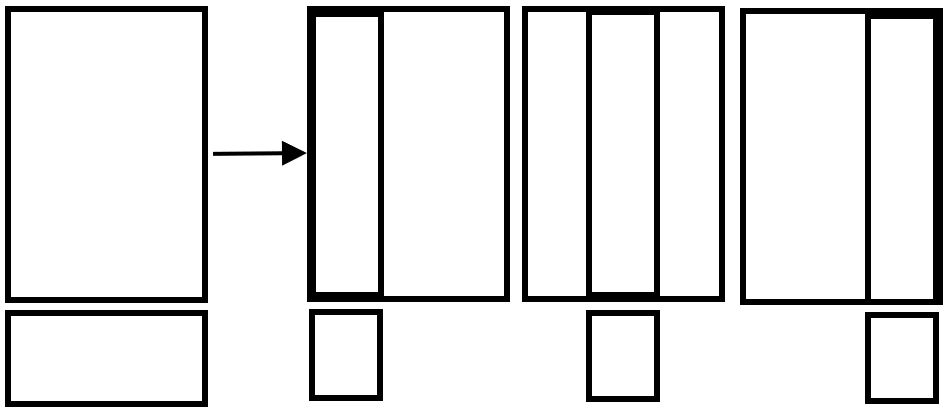}};  \node at (0.50,1.57) {\tiny Train}; \node at (0.46,0.91) {\tiny Test}; \\};
		\node [matrix, below right=of Resample] (A) {\node [block]  (Bagging) {  \includegraphics[width = 3cm]{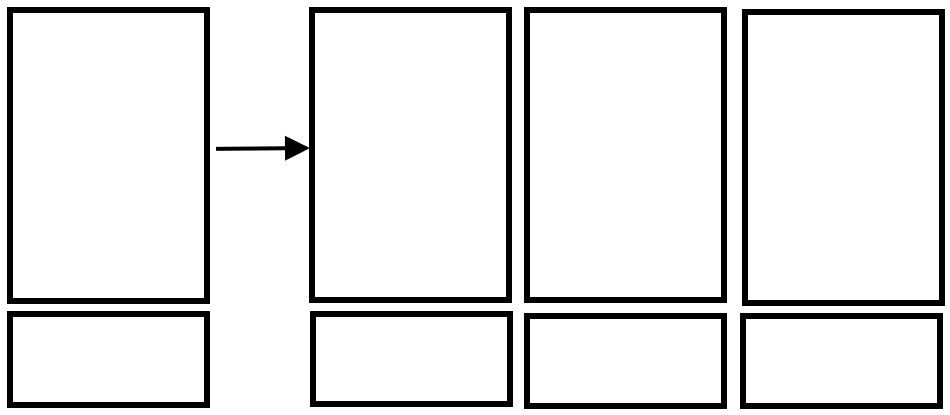}}; \node at (0.17,-0.4) {\tiny Train}; \node at (0.18,-1.09) {\tiny Test}; \\};
	\node [block1, below of = Resample,node distance = 5cm] (find) { Find the parameters and selected features based on Algorithm ({\color{blue}1}): SDFO-TR};
	\node [ align=center,below of= find,,node distance = 1cm] (init) {Initilize the ensemble model};	
	\node [blockPic, inner sep=0.5, ,outer sep=0.5,align=center,below of = init,node distance = 2.3cm] (ensemble) { \includegraphics[width = 2cm]{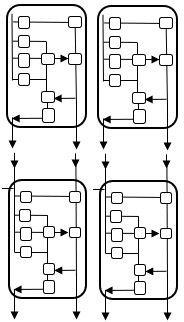}};
	\node [decision, below of = ensemble,node distance = 4cm] (Budget) { Function evaluation {\bf Left}?};
	\node [block1, , align=center, below of = Budget, node distance = 3.5cm] (no block) {
		  Calculate the statistics};
	\node [ align=center,right of= Budget,node distance = 5cm] (dummy) {};

\tikzset{blue dotted/.style={draw=black!100!white, line width=1pt,
		dash pattern=on 1pt off 4pt on 6pt off 4pt,
		inner sep=1cm, rectangle, rounded corners}};

\node (dotted box) [blue dotted, 
fit = (find) (init) (ensemble) (Budget)] {};

  \node at (dotted box.north) [below, inner sep=3mm] {\textbf{DFO}};	
  \node at (additive.north) [above, inner sep=3mm] {\textbf{Additive}};
  \node at (Bagging.north) [above, inner sep=3mm ] (Bagging1) {\textbf{Bagging}};    

\path [line] (Data) -- (Resample);
\path [line] (Resample) -- (additive);
\path [line] (Resample) -- (Bagging);
\path [line] (additive) -- (find);
\path [line] (Bagging) -- (find);
\draw (Budget) --  (dummy);
\path [line] (dummy) |- (find);
\path [line] (ensemble) -- (Budget);
\path [line] (Budget) -- node { No}(no block);
\draw (find) -- (init);
\path [line] (init) -- (ensemble);

	\end{tikzpicture}

%% file: archiveRevised.bbl
\begin{thebibliography}{54}
\providecommand{\natexlab}[1]{#1}
\providecommand{\url}[1]{\texttt{#1}}
\expandafter\ifx\csname urlstyle\endcsname\relax
  \providecommand{\doi}[1]{doi: #1}\else
  \providecommand{\doi}{doi: \begingroup \urlstyle{rm}\Url}\fi

\bibitem[Zhou et~al.(2011)Zhou, Shi, and Li]{zhou2011fine}
Junyi Zhou, Jing Shi, and Gong Li.
\newblock Fine tuning support vector machines for short-term wind speed
  forecasting.
\newblock \emph{Energy Conversion and Management}, 52\penalty0 (4):\penalty0
  1990--1998, 2011.

\bibitem[Li and Shi(2010)]{li2010comparing}
Gong Li and Jing Shi.
\newblock On comparing three artificial neural networks for wind speed
  forecasting.
\newblock \emph{Applied Energy}, 87\penalty0 (7):\penalty0 2313--2320, 2010.

\bibitem[Pirhooshyaran and Snyder(2019)]{pirhooshyaran2019multivariate}
Mohammad Pirhooshyaran and Lawrence~V Snyder.
\newblock Multivariate, multistep forecasting, reconstruction and feature
  selection of ocean waves via recurrent and sequence-to-sequence networks.
\newblock \emph{arXiv preprint arXiv:1906.00195}, 2019.

\bibitem[Davy et~al.(2018)Davy, Gnatiuk, Pettersson, and
  Bobylev]{davy2018climate}
Richard Davy, Natalia Gnatiuk, Lasse Pettersson, and Leonid Bobylev.
\newblock Climate change impacts on wind energy potential in the european
  domain with a focus on the black sea.
\newblock \emph{Renewable and Sustainable Energy Reviews}, 81:\penalty0
  1652--1659, 2018.

\bibitem[Conn et~al.(2009)Conn, Scheinberg, and Vicente]{conn2009introduction}
Andrew~R Conn, Katya Scheinberg, and Luis~N Vicente.
\newblock \emph{Introduction to derivative-free optimization}, volume~8.
\newblock Siam, 2009.

\bibitem[Breiman(1996)]{breiman1996bagging}
Leo Breiman.
\newblock Bagging predictors.
\newblock \emph{Machine learning}, 24\penalty0 (2):\penalty0 123--140, 1996.

\bibitem[Soman et~al.(2010)Soman, Zareipour, Malik, and
  Mandal]{soman2010review}
Saurabh~S Soman, Hamidreza Zareipour, Om~Malik, and Paras Mandal.
\newblock A review of wind power and wind speed forecasting methods with
  different time horizons.
\newblock In \emph{North American Power Symposium 2010}, pages 1--8. IEEE,
  2010.

\bibitem[Woon et~al.(2014)Woon, Aung, and Madnick]{woon2014data}
Wei~Lee Woon, Zeyar Aung, and Stuart Madnick.
\newblock Data analytics for renewable energy integration.
\newblock In \emph{Second ECML PKDD Workshop, DARE}. Springer, 2014.

\bibitem[Madsen et~al.(2005{\natexlab{a}})Madsen, Nielsen, and
  Nielsen]{madsen2005tool}
Henrik Madsen, Henrik~Aalborg Nielsen, and Torben~Skov Nielsen.
\newblock A tool for predicting the wind power production of off-shore wind
  plants.
\newblock In \emph{Proceedings of the Copenhagen Offshore Wind Conference \&
  Exhibition}, 2005{\natexlab{a}}.

\bibitem[Madsen et~al.(2005{\natexlab{b}})Madsen, Pinson, Kariniotakis,
  Nielsen, and Nielsen]{madsen2005standardizing}
Henrik Madsen, Pierre Pinson, George Kariniotakis, Henrik~Aa Nielsen, and
  Torben~S Nielsen.
\newblock Standardizing the performance evaluation of short-term wind power
  prediction models.
\newblock \emph{Wind Engineering}, 29\penalty0 (6):\penalty0 475--489,
  2005{\natexlab{b}}.

\bibitem[Coiffier(2011)]{coiffier2011fundamentals}
Jean Coiffier.
\newblock \emph{Fundamentals of numerical weather prediction}.
\newblock Cambridge University Press, 2011.

\bibitem[Kavasseri and Seetharaman(2009)]{kavasseri2009day}
Rajesh~G Kavasseri and Krithika Seetharaman.
\newblock Day-ahead wind speed forecasting using f-arima models.
\newblock \emph{Renewable Energy}, 34\penalty0 (5):\penalty0 1388--1393, 2009.

\bibitem[Singh et~al.(2019)Singh, Mohapatra, et~al.]{singh2019repeated}
SN~Singh, Abheejeet Mohapatra, et~al.
\newblock Repeated wavelet transform based arima model for very short-term wind
  speed forecasting.
\newblock \emph{Renewable Energy}, 2019.

\bibitem[Tian et~al.(2018)Tian, Fu, Ling, Wei, Liu, and Li]{tian2018wind}
Shuxin Tian, Yang Fu, Ping Ling, Shurong Wei, Shu Liu, and Kunpeng Li.
\newblock Wind power forecasting based on arima-lgarch model.
\newblock In \emph{2018 International Conference on Power System Technology
  (POWERCON)}, pages 1285--1289. IEEE, 2018.

\bibitem[Khosravi et~al.(2018)Khosravi, Koury, Machado, and
  Pabon]{khosravi2018prediction}
A~Khosravi, RNN Koury, L~Machado, and JJG Pabon.
\newblock Prediction of wind speed and wind direction using artificial neural
  network, support vector regression and adaptive neuro-fuzzy inference system.
\newblock \emph{Sustainable Energy Technologies and Assessments}, 25:\penalty0
  146--160, 2018.

\bibitem[Moreno and dos Santos~Coelho(2018)]{moreno2018wind}
Sinvaldo~Rodrigues Moreno and Leandro dos Santos~Coelho.
\newblock Wind speed forecasting approach based on singular spectrum analysis
  and adaptive neuro fuzzy inference system.
\newblock \emph{Renewable energy}, 126:\penalty0 736--754, 2018.

\bibitem[Chitsazan et~al.(2019)Chitsazan, Fadali, and
  Trzynadlowski]{chitsazan2019wind}
Mohammad~Amin Chitsazan, M~Sami Fadali, and Andrzej~M Trzynadlowski.
\newblock Wind speed and wind direction forecasting using echo state network
  with nonlinear functions.
\newblock \emph{Renewable energy}, 131:\penalty0 879--889, 2019.

\bibitem[Sutskever et~al.(2014)Sutskever, Vinyals, and
  Le]{sutskever2014sequence}
Ilya Sutskever, Oriol Vinyals, and Quoc~V Le.
\newblock Sequence to sequence learning with neural networks.
\newblock In \emph{Advances in neural information processing systems}, pages
  3104--3112, 2014.

\bibitem[Xiong et~al.(2018)Xiong, Wu, Alleva, Droppo, Huang, and
  Stolcke]{xiong2018microsoft}
Wayne Xiong, Lingfeng Wu, Fil Alleva, Jasha Droppo, Xuedong Huang, and Andreas
  Stolcke.
\newblock The microsoft 2017 conversational speech recognition system.
\newblock In \emph{2018 IEEE International Conference on Acoustics, Speech and
  Signal Processing (ICASSP)}, pages 5934--5938. IEEE, 2018.

\bibitem[Zhang et~al.(2018)Zhang, Yin, Zhang, Liu, and
  Bengio]{zhang2018drawing}
Xu-Yao Zhang, Fei Yin, Yan-Ming Zhang, Cheng-Lin Liu, and Yoshua Bengio.
\newblock Drawing and recognizing chinese characters with recurrent neural
  network.
\newblock \emph{IEEE transactions on pattern analysis and machine
  intelligence}, 40\penalty0 (4):\penalty0 849--862, 2018.

\bibitem[Mobiny(2018)]{mobiny2018text}
Aryan Mobiny.
\newblock Text-independent speaker verification using long short-term memory
  networks.
\newblock \emph{arXiv preprint arXiv:1805.00604}, 2018.

\bibitem[Lipton et~al.(2015)Lipton, Kale, Elkan, and
  Wetzel]{lipton2015learning}
Zachary~C Lipton, David~C Kale, Charles Elkan, and Randall Wetzel.
\newblock Learning to diagnose with lstm recurrent neural networks.
\newblock \emph{arXiv preprint arXiv:1511.03677}, 2015.

\bibitem[Mobiny et~al.(2017)Mobiny, Moulik, and Van~Nguyen]{mobiny2017lung}
Aryan Mobiny, Supratik Moulik, and Hien Van~Nguyen.
\newblock Lung cancer screening using adaptive memory-augmented recurrent
  networks.
\newblock \emph{arXiv preprint arXiv:1710.05719}, 2017.

\bibitem[Choi et~al.(2016)Choi, Fazekas, and Sandler]{choi2016text}
Keunwoo Choi, George Fazekas, and Mark Sandler.
\newblock Text-based lstm networks for automatic music composition.
\newblock \emph{arXiv preprint arXiv:1604.05358}, 2016.

\bibitem[Ghanbari and Scheinberg(2017)]{ghanbari2017black}
Hiva Ghanbari and Katya Scheinberg.
\newblock Black-box optimization in machine learning with trust region based
  derivative free algorithm.
\newblock \emph{arXiv preprint arXiv:1703.06925}, 2017.

\bibitem[Silva et~al.(2013)Silva, Morais, Sousa, and Vale]{silva2013energy}
Marco Silva, Hugo Morais, Tiago Sousa, and Zita Vale.
\newblock Energy resources management in three distinct time horizons
  considering a large variation in wind power.
\newblock \emph{EWEA Annual Event 2013 (EWEA 2013)}, 2013.

\bibitem[Pinto et~al.(2014)Pinto, Ramos, Sousa, and Vale]{pinto2014short}
Tiago Pinto, S{\'e}rgio Ramos, Tiago~M Sousa, and Zita Vale.
\newblock Short-term wind speed forecasting using support vector machines.
\newblock In \emph{2014 IEEE Symposium on Computational Intelligence in Dynamic
  and Uncertain Environments (CIDUE)}, pages 40--46. IEEE, 2014.

\bibitem[Ramos et~al.(2013)Ramos, Soares, Pinto, and Vale]{ramos2013short}
S{\'e}rgio Ramos, Jo{\~a}o Soares, Tiago Pinto, and Zita Vale.
\newblock Short-term wind forecasting to support virtual power player
  operation.
\newblock \emph{EWEA Annual Event 2013 (EWEA 2013)}, 2013.

\bibitem[B{\"u}hlmann et~al.(2002)B{\"u}hlmann, Yu,
  et~al.]{buhlmann2002analyzing}
Peter B{\"u}hlmann, Bin Yu, et~al.
\newblock Analyzing bagging.
\newblock \emph{The Annals of Statistics}, 30\penalty0 (4):\penalty0 927--961,
  2002.

\bibitem[Conn et~al.(1997)Conn, Scheinberg, and Toint]{conn1997convergence}
Andrew~R Conn, Katya Scheinberg, and Ph~L Toint.
\newblock On the convergence of derivative-free methods for unconstrained
  optimization.
\newblock \emph{Approximation theory and optimization: tributes to MJD Powell},
  pages 83--108, 1997.

\bibitem[Rios and Sahinidis(2013)]{rios2013derivative}
Luis~Miguel Rios and Nikolaos~V Sahinidis.
\newblock Derivative-free optimization: a review of algorithms and comparison
  of software implementations.
\newblock \emph{Journal of Global Optimization}, 56\penalty0 (3):\penalty0
  1247--1293, 2013.

\bibitem[Li et~al.(2014)Li, Zhang, Chen, and Smola]{li2014efficient}
Mu~Li, Tong Zhang, Yuqiang Chen, and Alexander~J Smola.
\newblock Efficient mini-batch training for stochastic optimization.
\newblock In \emph{Proceedings of the 20th ACM SIGKDD international conference
  on Knowledge discovery and data mining}, pages 661--670. ACM, 2014.

\bibitem[Mor{\'e} and Wild(2009)]{more2009benchmarking}
Jorge~J Mor{\'e} and Stefan~M Wild.
\newblock Benchmarking derivative-free optimization algorithms.
\newblock \emph{SIAM Journal on Optimization}, 20\penalty0 (1):\penalty0
  172--191, 2009.

\bibitem[Box et~al.(2015)Box, Jenkins, Reinsel, and Ljung]{box2015time}
George~EP Box, Gwilym~M Jenkins, Gregory~C Reinsel, and Greta~M Ljung.
\newblock \emph{Time series analysis: forecasting and control}.
\newblock John Wiley \& Sons, 2015.

\bibitem[Pirhooshyaran and Snyder(2017)]{pirhooshyaran2017optimization}
Mohammad Pirhooshyaran and Lawrence~V Snyder.
\newblock Optimization of inventory and distribution for hip and knee joint
  replacements via multistage stochastic programming.
\newblock In \emph{Modeling and optimization: Theory and applications}, pages
  139--155. Springer, 2017.

\bibitem[Pirhooshyaran and Niaki(2015)]{pirhooshyaran2015double}
Mohammad Pirhooshyaran and Seyed Taghi~Akhavan Niaki.
\newblock A double-max mewma scheme for simultaneous monitoring and fault
  isolation of multivariate multistage auto-correlated processes based on novel
  reduced-dimension statistics.
\newblock \emph{Journal of Process Control}, 29:\penalty0 11--22, 2015.

\bibitem[Broomhead and Lowe(1988)]{broomhead1988radial}
David~S Broomhead and David Lowe.
\newblock Radial basis functions, multi-variable functional interpolation and
  adaptive networks.
\newblock Technical report, Royal Signals and Radar Establishment Malvern
  (United Kingdom), 1988.

\bibitem[Pao(1989)]{pao1989adaptive}
Yohhan Pao.
\newblock Adaptive pattern recognition and neural networks.
\newblock 1989.

\bibitem[Huang et~al.(2006)Huang, Zhu, and Siew]{huang2006extreme}
Guang-Bin Huang, Qin-Yu Zhu, and Chee-Kheong Siew.
\newblock Extreme learning machine: theory and applications.
\newblock \emph{Neurocomputing}, 70\penalty0 (1-3):\penalty0 489--501, 2006.

\bibitem[Miche et~al.(2010)Miche, Sorjamaa, Bas, Simula, Jutten, and
  Lendasse]{miche2010op}
Yoan Miche, Antti Sorjamaa, Patrick Bas, Olli Simula, Christian Jutten, and
  Amaury Lendasse.
\newblock Op-elm: optimally pruned extreme learning machine.
\newblock \emph{IEEE transactions on neural networks}, 21\penalty0
  (1):\penalty0 158--162, 2010.

\bibitem[Park and Kim(2017)]{park2017online}
Jin-Man Park and Jong-Hwan Kim.
\newblock Online recurrent extreme learning machine and its application to
  time-series prediction.
\newblock In \emph{2017 International Joint Conference on Neural Networks
  (IJCNN)}, pages 1983--1990. IEEE, 2017.

\bibitem[Wong et~al.(2014)Wong, Vong, Gao, and Wong]{wong2014adaptive}
Pak~Kin Wong, Chi~Man Vong, Xiang~Hui Gao, and Ka~In Wong.
\newblock Adaptive control using fully online sequential-extreme learning
  machine and a case study on engine air-fuel ratio regulation.
\newblock \emph{Mathematical Problems in Engineering}, 2014, 2014.

\bibitem[Drucker et~al.(1997)Drucker, Burges, Kaufman, Smola, and
  Vapnik]{drucker1997support}
Harris Drucker, Christopher~JC Burges, Linda Kaufman, Alex~J Smola, and
  Vladimir Vapnik.
\newblock Support vector regression machines.
\newblock In \emph{Advances in neural information processing systems}, pages
  155--161, 1997.

\bibitem[Sadrfaridpour et~al.(2016)Sadrfaridpour, Jeereddy, Kennedy, Luckow,
  Razzaghi, and Safro]{sadrfaridpour2016algebraic}
Ehsan Sadrfaridpour, Sandeep Jeereddy, Ken Kennedy, Andre Luckow, Talayeh
  Razzaghi, and Ilya Safro.
\newblock Algebraic multigrid support vector machines.
\newblock \emph{arXiv preprint arXiv:1611.05487}, 2016.

\bibitem[Azadeh et~al.(2015)Azadeh, Boskabadi, and Pashapour]{azadeh2015unique}
Ali Azadeh, Azam Boskabadi, and Shima Pashapour.
\newblock A unique support vector regression for improved modelling and
  forecasting of short-term gasoline consumption in railway systems.
\newblock \emph{International Journal of Services and Operations Management},
  21\penalty0 (2):\penalty0 217--237, 2015.

\bibitem[Chang and Lin(2011)]{CC01a}
Chih-Chung Chang and Chih-Jen Lin.
\newblock {LIBSVM}: A library for support vector machines.
\newblock \emph{ACM Transactions on Intelligent Systems and Technology},
  2:\penalty0 27:1--27:27, 2011.
\newblock Software available at \url{http://www.csie.ntu.edu.tw/~cjlin/libsvm}.

\bibitem[Breiman(2001)]{breiman2001random}
Leo Breiman.
\newblock Random forests.
\newblock \emph{Machine learning}, 45\penalty0 (1):\penalty0 5--32, 2001.

\bibitem[Notton et~al.(2018)Notton, Nivet, Voyant, Paoli, Darras, Motte, and
  Fouilloy]{notton2018intermittent}
Gilles Notton, Marie-Laure Nivet, Cyril Voyant, Christophe Paoli, Christophe
  Darras, Fabrice Motte, and Alexis Fouilloy.
\newblock Intermittent and stochastic character of renewable energy sources:
  Consequences, cost of intermittence and benefit of forecasting.
\newblock \emph{Renewable and Sustainable Energy Reviews}, 87:\penalty0
  96--105, 2018.

\bibitem[Kalogirou(2013)]{kalogirou2013solar}
Soteris~A Kalogirou.
\newblock \emph{Solar energy engineering: processes and systems}.
\newblock Academic Press, 2013.

\bibitem[Ameen et~al.(2019)Ameen, Balzter, Jarvis, and
  Wheeler]{ameen2019modelling}
Bikhtiyar Ameen, Heiko Balzter, Claire Jarvis, and James Wheeler.
\newblock Modelling hourly global horizontal irradiance from satellite-derived
  datasets and climate variables as new inputs with artificial neural networks.
\newblock \emph{Energies}, 12\penalty0 (1):\penalty0 148, 2019.

\bibitem[Sengupta et~al.(2018)Sengupta, Xie, Lopez, Habte, Maclaurin, and
  Shelby]{sengupta2018national}
Manajit Sengupta, Yu~Xie, Anthony Lopez, Aron Habte, Galen Maclaurin, and James
  Shelby.
\newblock The national solar radiation data base (nsrdb).
\newblock \emph{Renewable and Sustainable Energy Reviews}, 89:\penalty0 51--60,
  2018.

\bibitem[Lehmann et~al.(2017)Lehmann, Karimpour, Goudey, Jacobson, and
  Alam]{lehmann2017ocean}
Marcus Lehmann, Farid Karimpour, Clifford~A Goudey, Paul~T Jacobson, and
  Mohammad-Reza Alam.
\newblock Ocean wave energy in the united states: Current status and future
  perspectives.
\newblock \emph{Renewable and Sustainable Energy Reviews}, 74:\penalty0
  1300--1313, 2017.

\bibitem[Aderinto and Li(2018)]{aderinto2018ocean}
Tunde Aderinto and Hua Li.
\newblock Ocean wave energy converters: Status and challenges.
\newblock \emph{Energies}, 11\penalty0 (5):\penalty0 1250, 2018.

\bibitem[Greff et~al.(2016)Greff, Srivastava, Koutn{\'\i}k, Steunebrink, and
  Schmidhuber]{greff2016lstm}
Klaus Greff, Rupesh~K Srivastava, Jan Koutn{\'\i}k, Bas~R Steunebrink, and
  J{\"u}rgen Schmidhuber.
\newblock Lstm: A search space odyssey.
\newblock \emph{IEEE transactions on neural networks and learning systems},
  28\penalty0 (10):\penalty0 2222--2232, 2016.

\end{thebibliography}
